%% file: mainpaper.tex
\documentclass[runningheads]{llncs}
\usepackage{amssymb}
\usepackage{latexsym}
\usepackage{amsmath}
\PassOptionsToPackage{hyphens}{url}
\usepackage{hyperref}
\usepackage[T1]{fontenc}
\usepackage{float}
\usepackage{multirow}
\usepackage{threeparttable}
\usepackage{booktabs}
\usepackage{xspace}
\usepackage{bbding} %重要：首先在导言区调用bbding包
\usepackage{changepage}   % Required for the adjustwidth environment
\usepackage{pifont} % For checkmark and X mark
\usepackage{geometry} %geometry命令调整页边距
\usepackage{graphicx}
\begin{document}
\title{A Survey on Visual Mamba}
\author{Hanwei Zhang\inst{1,4,5} \and
Ying Zhu\inst{2} \and
Dan Wang\inst{2}\and
Lijun Zhang\inst{1} \and
Tianxiang Chen\inst{3}\and
Zi Ye\inst{4}}
\authorrunning{H. Zhang et al.}
\institute{Automotive Software Innovation Center, Chongqing 401331, China\\
\email{zhang@depend.uni-saarland.de}
\\ \email{zhanglj@ios.ac.cn}
\and Hangzhou Institute for Advanced Study, University of Chinese Academy of Sciences, Hangzhou 310024, China\\
\email{zhuying23@mails.ucas.ac.cn}
\\ \email{wangdan233@mails.ucas.ac.cn}\and
University of Science and Technology of China\\
\email{txchen@mail.ustc.edu.cn}\and
Institute of Intelligent Software, Guangzhou, China\\
\email{yezi1022@gmail.com}\and
Saarland University, Germany } 
\maketitle              % typeset the header of the contribution
\begin{abstract}

State space models (SSMs) with selection mechanisms and hardware-aware architectures, namely Mamba, have recently demonstrated significant promise in long-sequence modeling. Since the self-attention mechanism in transformers has quadratic complexity with image size and increasing computational demands, the researchers are now exploring how to adapt Mamba for computer vision tasks. This paper is the first comprehensive survey aiming to provide an in-depth analysis of Mamba models in the field of computer vision. It begins by exploring the foundational concepts contributing to Mamba's success, including the state space model framework, selection mechanisms, and hardware-aware design. Next, we review these vision mamba models by categorizing them into foundational ones and enhancing them with techniques such as convolution, recurrence, and attention to improve their sophistication. We further delve into the widespread applications of Mamba in vision tasks, which include their use as a backbone in various levels of vision processing. This encompasses general visual tasks, Medical visual tasks (e.g., 2D / 3D segmentation, classification, and image registration, etc.), and Remote Sensing visual tasks. We specially introduce general visual tasks from two levels: High/Mid-level vision (e.g., Object detection, Segmentation, Video classification, etc.) and Low-level vision (e.g., Image super-resolution, Image restoration, Visual generation, etc.).  We hope this endeavor will spark additional interest within the community to address current challenges and further apply Mamba models in computer vision.

\end{abstract}

\input{tex/def}
\input{tex/intro}

\input{tex/FormulationMamba}
\input{tex/mamba}

\input{tex/ApplicationTask}
\input{tex/conclu}

\begin{credits}
\subsubsection{\ackname} No acknowledgments.

\subsubsection{\discintname}
The authors have no competing interests to declare
relevant to this article's content.
\end{credits}
\bibliographystyle{unsrt}
\bibliography{refbib}

\end{document}

%% file: tex/def.tex
\newcommand{\head}[1]{{\smallskip\noindent\textbf{#1}}}
\newcommand{\alert}[1]{{\color{red}{#1}}}
\newcommand{\sm}{\scriptsize}
\newcommand{\eq}[1]{(\ref{eq:#1})}

\newcommand{\Th}[1]{\textsc{#1}}
\newcommand{\mr}[2]{\multirow{#1}{*}{#2}}
\newcommand{\mc}[2]{\multicolumn{#1}{c}{#2}}
\newcommand{\tb}[1]{\textbf{#1}}
\newcommand{\ch}{\checkmark}

\newcommand{\red}[1]{{\textcolor{red}{#1}}}
\newcommand{\blue}[1]{{\textcolor{blue}{#1}}}
\newcommand{\green}[1]{{\textcolor{green}{#1}}}
\newcommand{\gray}[1]{{\textcolor{gray}{#1}}}

\newcommand{\citeme}[1]{\red{[XX]}}
\newcommand{\refme}[1]{\red{(XX)}}

\newcommand{\figh}[2][1]{\includegraphics[height=#1\linewidth]{fig/#2}}
\newcommand{\figa}[2][1]{\includegraphics[width=#1]{fig/#2}}
\newcommand{\figah}[2][1]{\includegraphics[height=#1]{fig/#2}}

\newcommand{\tran}{^\top}
\newcommand{\mtran}{^{-\top}}
\newcommand{\zcol}{\mathbf{0}}
\newcommand{\zrow}{\zcol\tran}

\newcommand{\ind}{\mathbbm{1}}
\newcommand{\expect}{\mathbb{E}}
\newcommand{\nat}{\mathbb{N}}
\newcommand{\zahl}{\mathbb{Z}}
\newcommand{\real}{\mathbb{R}}
\newcommand{\proj}{\mathbb{P}}
\newcommand{\prob}{\operatorname{P}}
\newcommand{\normal}{\mathcal{N}}

\newcommand{\mif}{\textrm{if}\ }
\newcommand{\other}{\textrm{otherwise}}
\newcommand{\minimize}{\textrm{minimize}\ }
\newcommand{\maximize}{\textrm{maximize}\ }

\newcommand{\id}{\operatorname{id}}
\newcommand{\const}{\operatorname{const}}
\newcommand{\sgn}{\operatorname{sgn}}
\newcommand{\var}{\operatorname{Var}}
\newcommand{\mean}{\operatorname{mean}}
\newcommand{\trace}{\operatorname{tr}}
\newcommand{\diag}{\operatorname{diag}}
\newcommand{\vect}{\operatorname{vec}}
\newcommand{\cov}{\operatorname{cov}}
\newcommand{\sign}{\operatorname{sign}}
\newcommand{\prj}{\operatorname{proj}}

\newcommand{\softmax}{\operatorname{softmax}}
\newcommand{\clip}{\operatorname{clip}}

\newcommand{\defn}{\mathrel{:=}}
\newcommand{\peq}{\mathrel{+\!=}}
\newcommand{\meq}{\mathrel{-\!=}}

\newcommand{\paren}[1]{\left({#1}\right)}
\newcommand{\mat}[1]{\left[{#1}\right]}
\newcommand{\set}[1]{\left\{{#1}\right\}}
\newcommand{\floor}[1]{\left\lfloor{#1}\right\rfloor}
\newcommand{\ceil}[1]{\left\lceil{#1}\right\rceil}
\newcommand{\inner}[1]{\left\langle{#1}\right\rangle}
\newcommand{\norm}[1]{\left\|{#1}\right\|}
\newcommand{\abs}[1]{\left|{#1}\right|}
\newcommand{\frob}[1]{\norm{#1}_F}
\newcommand{\card}[1]{\left|{#1}\right|\xspace}

\newcommand{\diff}{\mathrm{d}}
\newcommand{\der}[3][]{\frac{\diff^{#1}#2}{\diff#3^{#1}}}
\newcommand{\ider}[3][]{\diff^{#1}#2/\diff#3^{#1}}
\newcommand{\pder}[3][]{\frac{\partial^{#1}{#2}}{\partial{{#3}^{#1}}}}
\newcommand{\ipder}[3][]{\partial^{#1}{#2}/\partial{#3^{#1}}}
\newcommand{\dder}[3]{\frac{\partial^2{#1}}{\partial{#2}\partial{#3}}}

\newcommand{\wb}[1]{\overline{#1}}
\newcommand{\wt}[1]{\widetilde{#1}}

\def\xssp{\hspace{1pt}}
\def\ssp{\hspace{3pt}}
\def\msp{\hspace{5pt}}
\def\lsp{\hspace{12pt}}

\newcommand{\cA}{\mathcal{A}}
\newcommand{\cB}{\mathcal{B}}
\newcommand{\cC}{\mathcal{C}}
\newcommand{\cD}{\mathcal{D}}
\newcommand{\cE}{\mathcal{E}}
\newcommand{\cF}{\mathcal{F}}
\newcommand{\cG}{\mathcal{G}}
\newcommand{\cH}{\mathcal{H}}
\newcommand{\cI}{\mathcal{I}}
\newcommand{\cJ}{\mathcal{J}}
\newcommand{\cK}{\mathcal{K}}
\newcommand{\cL}{\mathcal{L}}
\newcommand{\cM}{\mathcal{M}}
\newcommand{\cN}{\mathcal{N}}
\newcommand{\cO}{\mathcal{O}}
\newcommand{\cP}{\mathcal{P}}
\newcommand{\cQ}{\mathcal{Q}}
\newcommand{\cR}{\mathcal{R}}
\newcommand{\cS}{\mathcal{S}}
\newcommand{\cT}{\mathcal{T}}
\newcommand{\cU}{\mathcal{U}}
\newcommand{\cV}{\mathcal{V}}
\newcommand{\cW}{\mathcal{W}}
\newcommand{\cX}{\mathcal{X}}
\newcommand{\cY}{\mathcal{Y}}
\newcommand{\cZ}{\mathcal{Z}}

\newcommand{\vA}{\mathbf{A}}
\newcommand{\vB}{\mathbf{B}}
\newcommand{\vC}{\mathbf{C}}
\newcommand{\vD}{\mathbf{D}}
\newcommand{\vE}{\mathbf{E}}
\newcommand{\vF}{\mathbf{F}}
\newcommand{\vG}{\mathbf{G}}
\newcommand{\vH}{\mathbf{H}}
\newcommand{\vI}{\mathbf{I}}
\newcommand{\vJ}{\mathbf{J}}
\newcommand{\vK}{\mathbf{K}}
\newcommand{\vL}{\mathbf{L}}
\newcommand{\vM}{\mathbf{M}}
\newcommand{\vN}{\mathbf{N}}
\newcommand{\vO}{\mathbf{O}}
\newcommand{\vP}{\mathbf{P}}
\newcommand{\vQ}{\mathbf{Q}}
\newcommand{\vR}{\mathbf{R}}
\newcommand{\vS}{\mathbf{S}}
\newcommand{\vT}{\mathbf{T}}
\newcommand{\vU}{\mathbf{U}}
\newcommand{\vV}{\mathbf{V}}
\newcommand{\vW}{\mathbf{W}}
\newcommand{\vX}{\mathbf{X}}
\newcommand{\vY}{\mathbf{Y}}
\newcommand{\vZ}{\mathbf{Z}}

\newcommand{\va}{\mathbf{a}}
\newcommand{\vb}{\mathbf{b}}
\newcommand{\vc}{\mathbf{c}}
\newcommand{\vd}{\mathbf{d}}
\newcommand{\ve}{\mathbf{e}}
\newcommand{\vf}{\mathbf{f}}
\newcommand{\vg}{\mathbf{g}}
\newcommand{\vh}{\mathbf{h}}
\newcommand{\vi}{\mathbf{i}}
\newcommand{\vj}{\mathbf{j}}
\newcommand{\vk}{\mathbf{k}}
\newcommand{\vl}{\mathbf{l}}
\newcommand{\vm}{\mathbf{m}}
\newcommand{\vn}{\mathbf{n}}
\newcommand{\vo}{\mathbf{o}}
\newcommand{\vp}{\mathbf{p}}
\newcommand{\vq}{\mathbf{q}}
\newcommand{\vr}{\mathbf{r}}
\newcommand{\Vs}{\mathbf{s}}
\newcommand{\vt}{\mathbf{t}}
\newcommand{\vu}{\mathbf{u}}
\newcommand{\vv}{\mathbf{v}}
\newcommand{\vw}{\mathbf{w}}
\newcommand{\vx}{\mathbf{x}}
\newcommand{\vy}{\mathbf{y}}
\newcommand{\vz}{\mathbf{z}}

\newcommand{\vone}{\mathbf{1}}
\newcommand{\vzero}{\mathbf{0}}

\newcommand{\valpha}{{\boldsymbol{\alpha}}}
\newcommand{\vbeta}{{\boldsymbol{\beta}}}
\newcommand{\vgamma}{{\boldsymbol{\gamma}}}
\newcommand{\vdelta}{{\boldsymbol{\delta}}}
\newcommand{\vepsilon}{{\boldsymbol{\epsilon}}}
\newcommand{\vzeta}{{\boldsymbol{\zeta}}}
\newcommand{\veta}{{\boldsymbol{\eta}}}
\newcommand{\vtheta}{{\boldsymbol{\theta}}}
\newcommand{\viota}{{\boldsymbol{\iota}}}
\newcommand{\vkappa}{{\boldsymbol{\kappa}}}
\newcommand{\vlambda}{{\boldsymbol{\lambda}}}
\newcommand{\vmu}{{\boldsymbol{\mu}}}
\newcommand{\vnu}{{\boldsymbol{\nu}}}
\newcommand{\vxi}{{\boldsymbol{\xi}}}
\newcommand{\vomikron}{{\boldsymbol{\omikron}}}
\newcommand{\vpi}{{\boldsymbol{\pi}}}
\newcommand{\vrho}{{\boldsymbol{\rho}}}
\newcommand{\vsigma}{{\boldsymbol{\sigma}}}
\newcommand{\vtau}{{\boldsymbol{\tau}}}
\newcommand{\vupsilon}{{\boldsymbol{\upsilon}}}
\newcommand{\vphi}{{\boldsymbol{\phi}}}
\newcommand{\vchi}{{\boldsymbol{\chi}}}
\newcommand{\vpsi}{{\boldsymbol{\psi}}}
\newcommand{\vomega}{{\boldsymbol{\omega}}}

\newcommand{\rLambda}{\mathrm{\Lambda}}
\newcommand{\rSigma}{\mathrm{\Sigma}}

\newcommand{\vLambda}{\bm{\rLambda}}
\newcommand{\vSigma}{\bm{\rSigma}}

\makeatletter
\newcommand*\bdot{\mathpalette\bdot@{.7}}
\newcommand*\bdot@[2]{\mathbin{\vcenter{\hbox{\scalebox{#2}{$\m@th#1\bullet$}}}}}
\makeatother

\makeatletter
\DeclareRobustCommand\onedot{\futurelet\@let@token\@onedot}
\def\@onedot{\ifx\@let@token.\else.\null\fi\xspace}

\def\eg{\emph{e.g}\onedot} \def\Eg{\emph{E.g}\onedot}
\def\ie{\emph{i.e}\onedot} \def\Ie{\emph{I.e}\onedot}
\def\cf{\emph{cf}\onedot} \def\Cf{\emph{Cf}\onedot}
\def\etc{\emph{etc}\onedot} \def\vs{\emph{vs}\onedot}
\def\wrt{w.r.t\onedot} \def\dof{d.o.f\onedot} \def\aka{a.k.a\onedot}
\def\etal{\emph{et al}\onedot}
\makeatother

%% file: tex/intro.tex
\section{Introduction}

Deep Neural Networks (DNNs) have demonstrated remarkable performance across various artificial intelligence (AI) tasks, with the fundamental architecture playing a crucial role in determining the model's capabilities. Traditional neural networks typically comprise Multi-Layer Perceptron (MLP) or Fully Connected (FC) layers~\cite{rosenblatt1957perceptron,rosenblatt1962principles}. Convolutional neural networks (CNNs)~\cite{lecun1998gradient,krizhevsky2012imagenet} introduce convolutional and pooling layers, which are particularly effective for processing shift-invariant data like images. Recurrent neural networks (RNNs)~\cite{hochreiter1997long,bahdanau2014neural} utilize recurrent cells to handle sequential or time series data. To address the issue of CNN, RNN, and GNN models only capturing local relationships, the Transformer~\cite{parikh2016decomposable,vaswani2017attention,devlin2018bert}, introduced in 2017, excels at learning long-distance feature representations. Transformers primarily rely on attention-based attention mechanisms, \eg self-attention and cross-attention, to extract intrinsic features and improve their representation capability. Pre-trained massive transformer-based models, such as GPT-3~\cite{brown2020language}, deliver robust performance across various NLP datasets, excelling in natural language understanding and generation tasks.  
The remarkable performance of transformer-based models has propelled their widespread adoption in vision applications. The core of transformer models is their exceptional skill in capturing long-range dependencies and maximizing the use of large datasets. The feature extraction module is the main component of vision transformer architectures. It processes data using a sequence of self-attention blocks, significantly improving its capacity to analyze images.

\begin{figure}[ht]
\centering
\begin{tabular}{cc}
  \includegraphics[width=0.5\textwidth]{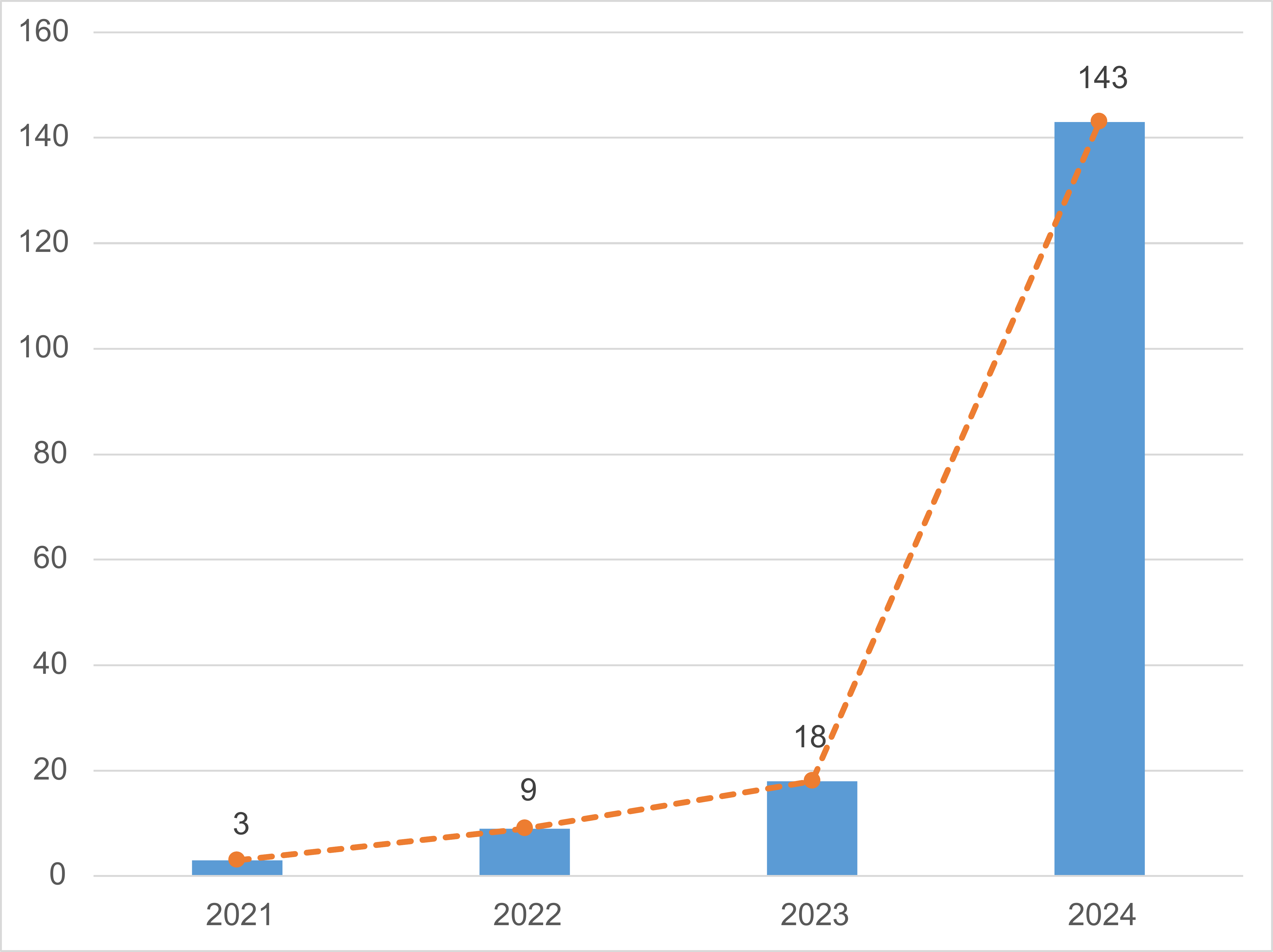}   &   \includegraphics[width=0.53\textwidth]{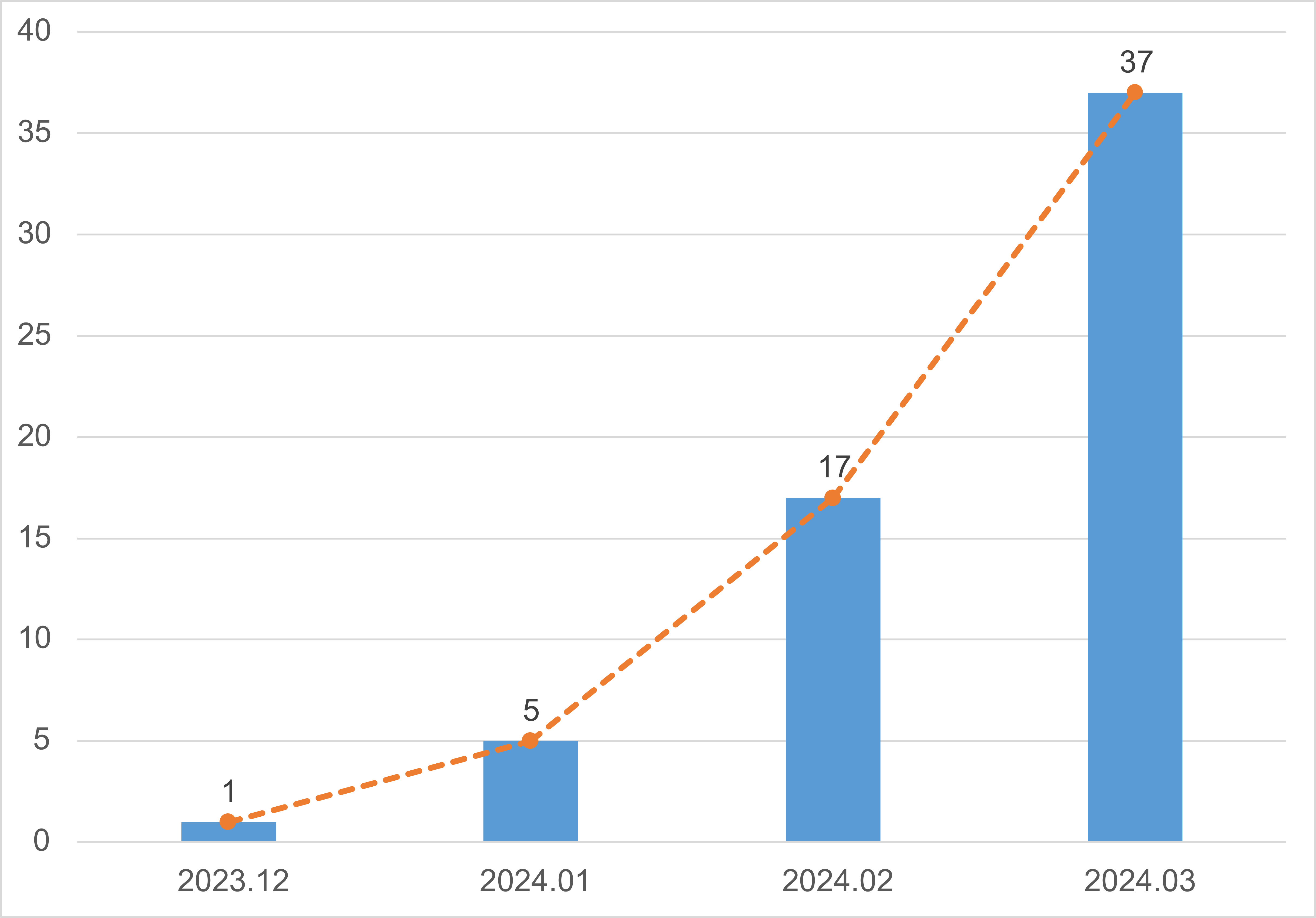} \\
    (a) SSM-based papers & (b) Mamba-based papers on vision
\end{tabular}

\caption{The number of SSMs and Mamba papers released to date(from year 2021 to year 2024.03).} \label{fig:SSMnumber}
\end{figure}

However, a primary obstacle for Transformers is the substantial computational demands of the self-attention mechanism, which increases quadratically with image resolution. The Softmax operation within the attention blocks further intensifies the computational demands, presenting significant challenges for implementing these models on edge and low-resource devices. Additionally, real-time computer vision systems utilizing transformer-based models must adhere to stringent low-latency standards to maintain a high-quality user experience. This scenario highlights the continuous evolution of new architectures to enhance performance, although this often comes with the trade-off of higher computational demands.
Many new models based on sparse attention mechanisms or innovative neural network paradigms have been proposed to reduce computational costs further while capturing long-range dependencies and maintaining high performance. State space models (SSMs) have emerged as a central focus among these developments. As shown in Fig.~\ref{fig:SSMnumber}(a), the number of publications related to SSMs demonstrates an explosive growth trend. Initially devised to simulate dynamic systems in areas such as control theory and computational neuroscience using state variables, SSMs predominantly describe linear invariant (or stable) systems when adapted for deep learning.

As SSMs have evolved, a new class of selective state space models, termed Mamba~\cite{gu2023mamba}. It has advanced the modeling of discrete data, such as text, with state-space models (SSMs) through two key improvements. Firstly, it features an input-dependent mechanism that adjusts SSM parameters dynamically, enhancing information filtering. Secondly, Mamba uses a hardware-aware algorithm that processes data linearly with sequence length, boosting computational speed on modern systems. Inspired by Mamba's achievements in language modeling, several initiatives are now aiming to adapt this success to the field of vision. Several studies have explored its integration with Mixture-of-Experts (MoE) techniques, as evidenced by works like Jamba~\cite{lieber2024jamba}, MoE-Mamba~\cite{pioro2024moe}, and BlackMamba~\cite{anthony2024blackmamba}, outperformed the state-of-the-art architecture Transformer-MoE with fewer training steps. As illustrated in Fig.~\ref{fig:SSMnumber}(b), since the release of Mamba in December 2023, the number of research papers focusing on Mamba in the vision domain has rapidly increased, reaching a peak in March 2024. This trend suggests that Mamba is emerging as a prominent research area in vision, potentially offering a viable alternative to Transformers. Therefore, A review of current related works is necessary and timely to provide a detailed overview of this new methodology in this evolving field.

Consequently, we present a comprehensive overview of how Mamba models are used in the vision domain. This paper aims to serve as a guide for researchers looking to delve deeper into this area. The critical contributions of our work include:

\begin{itemize}
  \item This survey paper is the first to provide a comprehensive review of the Mamba technique in the vision domain, explicitly focusing on analyzing the proposed strategies. 
 \item Expanding upon the Naive-based Mamba visual framework, we have investigated how Mamba's capabilities can be enhanced and combined with other architectures to achieve superior performance.
  \item We offer an in-depth exploration by organizing the literature based on various application tasks. We establish a taxonomy, identify advancements specific to each task, and offer insights on overcoming challenges.
\end{itemize}

The remainder of the survey is structured as follows: Section 2 examines the general and mathematical concepts underlying Mamba strategies. Section 3 discusses the naive Mamba visual models and how they integrate with other technologies to enhance performance, as proposed in recent years. Section 4 explores the application of Mamba technologies in addressing various computer vision tasks. Finally, Section 5 concludes the survey.

%% file: tex/FormulationMamba.tex
\section{Formulation of Mamba}

\begin{figure}[ht]
\includegraphics[width=\textwidth]{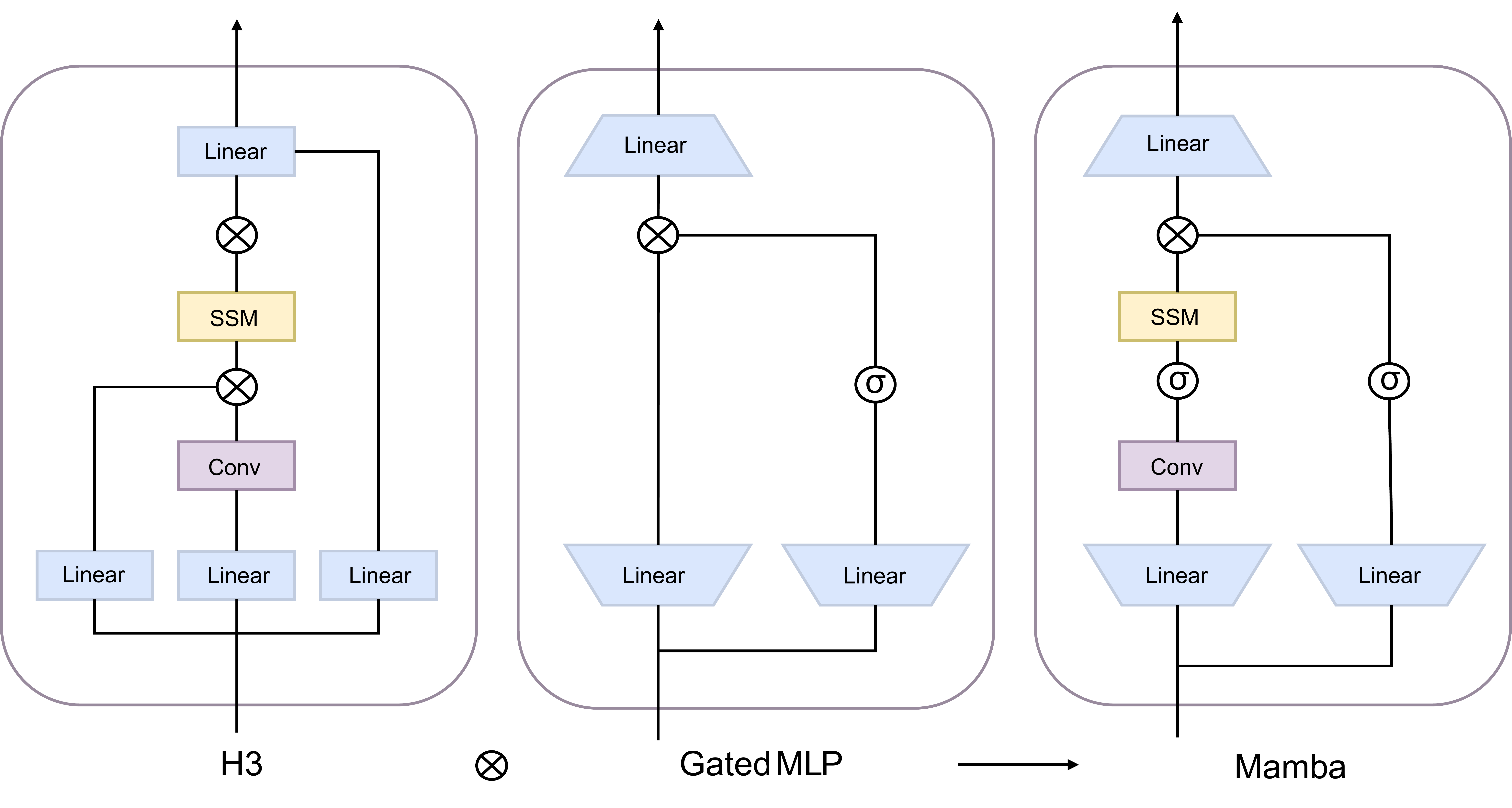}
\caption{Mamba Block~\cite{gu2023mamba}.} \label{fig:mamba-block}
\end{figure}

Mamba~\cite{gu2023mamba} is initially introduced in the domain of natural language processing. As shown in Fig.~\ref{fig:mamba-block}, the original Mamba Block integrates a Gated MLP into the State Space Model (SSM) architecture of H3~\cite{fu2022hungry}, utilizing an SSM sandwiched between two gated connections alongside a standard local convolution. For $\sigma$ SiLU~\cite{hendrycks2016gaussian} or Swish activation function~\cite{Ramachandran2017SwishAS} is used. The Mamba architecture consists of repeating of Mamba block interleaved with standard normalization and residual connections. An optional normalization layer (LayerNorm chosen by original Mamba) is used in a similar location as RetNet~\cite{sun2023retentive}.

\subsection{State Space Models (SSMs)}
Consider a structured state space model (SSM) that maps one-dimensional sequence $x(t) \in \real^L$ to $ y(t) \in \real^L$ through a hidden state $h(t) \in \real^N$. With the evolution parameter $\vA \in \real^{N \times N}$ and the projection parameters $\vB \in \real^{N \times 1}$, $\vC \in \real^{1 \times N}$, such a model is formulated as linear ordinary differential equations
\begin{align}
\begin{split}
    h'(t) = &~ \vA h(t) + \vB x(t), \\
    y(t) = &~ \vC h(t).      
\end{split}
\label{eq:ssms}
\end{align}

\paragraph{Discretization.} To adapt for deep learning, State Space Models (SSMs), as continuous-time models, are discretized with a Zero-Order Hold(ZOH) assumption. Thus, the continuous-time parameters $\vA, \vB$ are transform to their discretized counterparts $\overline{\vA}, \overline{\vB}$ with a timescale parameter $\Delta$ according to
\begin{align}
\begin{split}
    \overline{\vA} = &~\exp (\Delta \vA),\\
    \overline{\vB} = &~ (\Delta \vA)^{-1}(\exp(\Delta \vA)-\vI) \cdot \Delta \vB.
\end{split}
\label{eq:dis-A-B}
\end{align}
Thus, (\ref{eq:ssms}) can be rewritten as
\begin{align}
\begin{split}
    h_t =&~ \overline{\vA} h_{t-1} + \overline{\vB} x_t, \\
    y_t = &~ \vC h_t.
\end{split}
\label{eq:dis-ssms}
\end{align}
To enhance computational efficiency and scalability, the iterative process in (\ref{eq:dis-ssms}) can be synthesized through a global convolution
\begin{align}
\begin{split}
    \overline{\vK} =&~ (\vC\overline{\vB},\vC\overline{\vA}\overline{\vB},\cdots, \overline{\vA}^{L-1}\overline{\vB}),\\
    \vy = &~ \vx * \overline{\vK},
\end{split}
\label{eq:conv-ssms}
\end{align}
where $L$ is the length of the input sequence $\vx$, $\overline{\vK} \in \real^L$ serves as the kernel of the SSMs and $*$ represents the convolution operation.

\paragraph{Architectures.}
SSMs often serve as independent sequence transformations that can be integrated into end-to-end neural network architectures. Here we introduce several fundamental architectures. Linear attention~\cite{katharopoulos2020transformers} approximates self-attention with a recurrence mechanism as a simplified form of linear SSM.
H3~\cite{fu2022hungry}, as illustrated in Fig.~\ref{fig:mamba-block}, places an SSM between two gated connections and inserts a standard local convolution before it.
Following H3, Hyena~\cite{poli2023hyena}, replaces the SSM layer with an MLP-parameterized global convolution~\cite{romero2021ckconv}.
RetNet~\cite{sun2307retentive} introduces an extra gate and uses simpler SSM. RetNet enables an alternative parallelizable computation path and employs a variant of multi-head attention (MHA) instead of convolutions.
Inspired by attention-free Transformer~\cite{zhai2021attention}, the recent RNN design RWKV~\cite{peng2023rwkv}, can be interpreted as the ratio of two SSMs due to its primary "WKV" mechanism involving Linear Time Invariance (LTI) recurrences.

\paragraph{Selective State Space Models.}
Traditional SSMs demonstrated linear time complexity but their representativity of sequence context is inherently limited by time-invariant parameterization. To overcome this constraint, Selective State Space Models introduce \emph{selective scan} for interactions among sequential states with 
\begin{align}
\begin{split}
    \vB =&~ S_\vB(\vx), \\
    \vC =&~ S_\vC(\vx), \\
    \Delta =&~ \tau_\Delta( \Delta + S_\Delta(\vx)),
\end{split}
\label{eq:s6}
\end{align}
before (\ref{eq:dis-A-B},\ref{eq:dis-ssms}),
so that parameters $\vB \in \real^{B\times L \times N}$, $ \vC^{B\times L \times N}$ and $\Delta^{B\times L \times D}$ are dependent on the input sequence $\vx \in \real^{B\times L \times D}$, where $B$ represents the batch size, and $D$ represents number of channels. Normally, $S_B$ and $S_C$ are linear parameterized projections to dimension $N$, \ie $Linear_N(\cdot)$, while $S_\Delta(\vx) = Broadcast_D(Linear_1(\vx))$ and $\tau_\Delta = softplus$. The choice of $S_\Delta$ and $\tau_\Delta$ is due to a connection to RNN gating mechanisms explained later.

\subsection{Other Key Concepts in Mamba}

\paragraph{Selection Mechanism.} 

The connection between RNN gating and the discretization of continuous-time systems is well-established~\cite{tallec2018can}. The classical gating mechanism of RNNs is an instance of the selection mechanism for SSMs. When $N=1, \vA = -1, \vB = 1, S_\Delta = Linear(\vx)$ and $\tau_\Delta = softplus$, then the selective SSM recurrence takes the form
\begin{align}
    \begin{split}
        g_t =&~ \sigma(Linear(x(t)))\\
        h_t =&~ (1-g_t)h_{t-1} + g_t x_t.
    \end{split}
\end{align}

\paragraph{Scan.} The selection mechanism is devised to address the constraints of Linear Time Invariance (LTI) models. However, it reintroduces the computation issue associated with SSMs. To enhance GPU utilization and efficiently materialize the state $h$ within the memory hierarchy, hardware-aware state expansion is enabled by selective scan. 
By incorporating kernel fusion and recomputation with parallel scan, the fused selective scan layer effectively reduces the amount of memory I/O operations, leading to a significant acceleration compared to conventional implementations.

% \begin{figure}
% \includegraphics[width=\textwidth]{figures/covn-ssm-lstm.png}
% \caption{The connection between continous-time SSM and recurrent and convolutional model.(image from \cite{gu2021combining})} \label{fig:ssm-relation-others}
% \end{figure}

\paragraph{Discussion.}
Compared to RNNs and LSTMs, which struggle with vanishing gradients and long-range dependencies, Mamba offers efficient computation and memory utilization. While transformers excel in batch processing and handling long-range dependencies through attention mechanisms, they incur high computational costs, especially during inference. Mamba introduces a selective state space model, incorporating input-dependent matrices to enhance adaptability while maintaining the computational advantages of traditional SSMs. Mamba bridges the gap between traditional SSMs and modern neural network architectures by offering a selective dependency mechanism, optimal GPU memory utilization, and linear scalability with context length, thus providing a promising solution for various sequential data processing tasks.

%% file: tex/mamba.tex
\section{Mamba for Vision}

The original Mamba block is designed for one-dimensional sequences, yet vision-related tasks require processing multi-dimensional inputs like images, videos, and 3D representations. Consequently, to adapt Mamba for these tasks, enhancements to the scanning mechanism and architecture of the Mamba block are crucial to effectively handle multi-dimensional inputs. 

In this section, we present efforts aimed at enabling Mamba to tackle vision-related tasks while enhancing its efficiency and performance. Initially, we delve into two foundational works: Vision Mamba~\cite{zhu2024vision} and VMamaba~\cite{liu2024vmamba}. These works introduced the ViM block and VSS block, respectively, serving as the foundation for subsequent research endeavors. Subsequently, we explore additional works focused on refining the Mamba architecture as a backbone for vision-related tasks. Lastly, we discuss the work of integrating Mamba with other architectures such as convolution, recurrence, and attention.

\subsection{Visual Mamba Block.}

Drawing inspiration from the visual transformer architecture, it seems natural to preserve the framework of the transformer model while substituting the attention block with a Mamba block, while keeping the rest of the process intact. At the crux of the matter lies the adaptation of the Mamba block to vision-related tasks. Nearly simultaneously, Vision Mamba and VMamba present their respective solutions: the ViM block and the VSS block.

\begin{figure}[ht]
\begin{tabular}{ccc}
\includegraphics[width=0.4\textwidth]{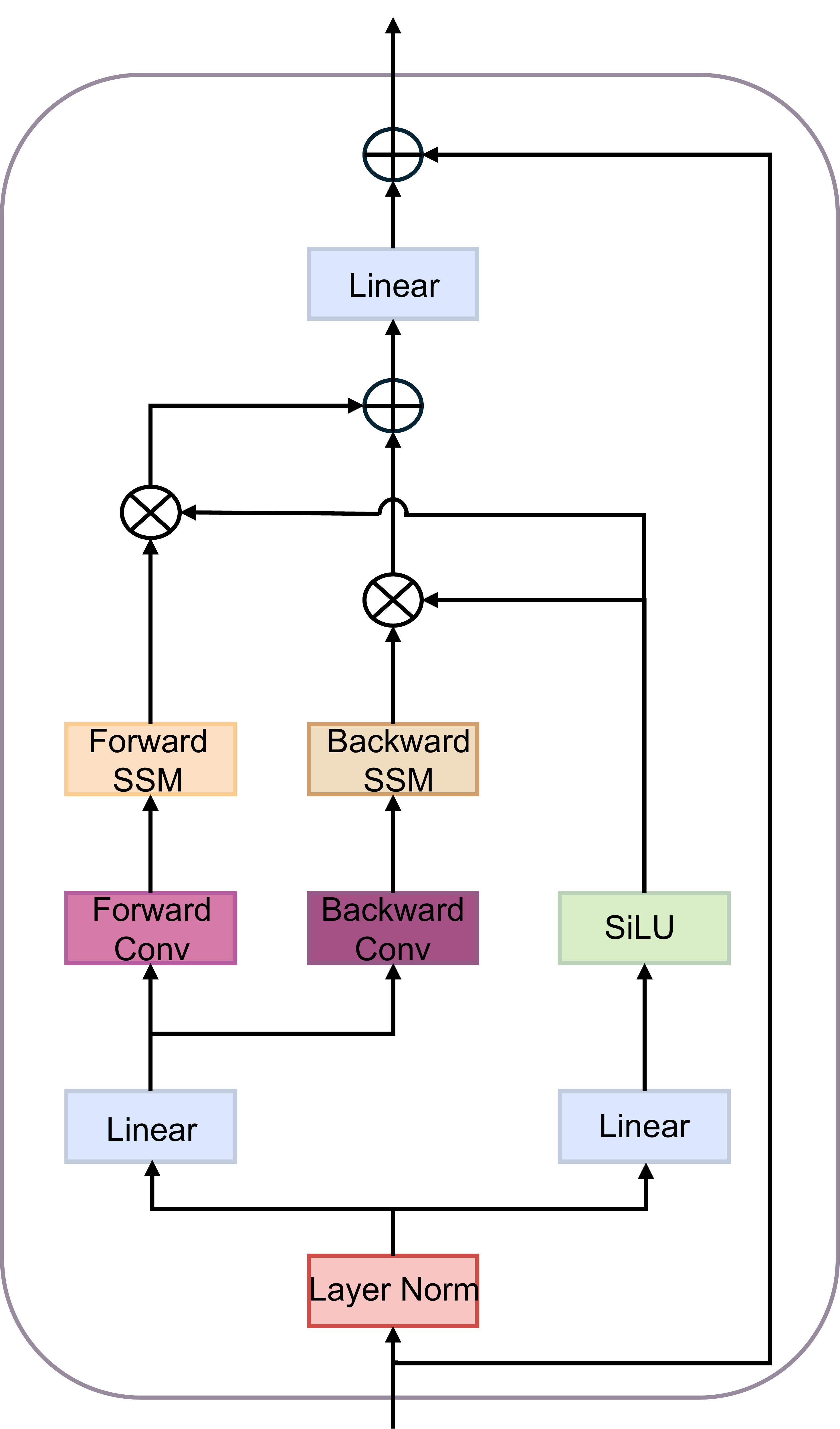} &
\includegraphics[width=0.4\textwidth]{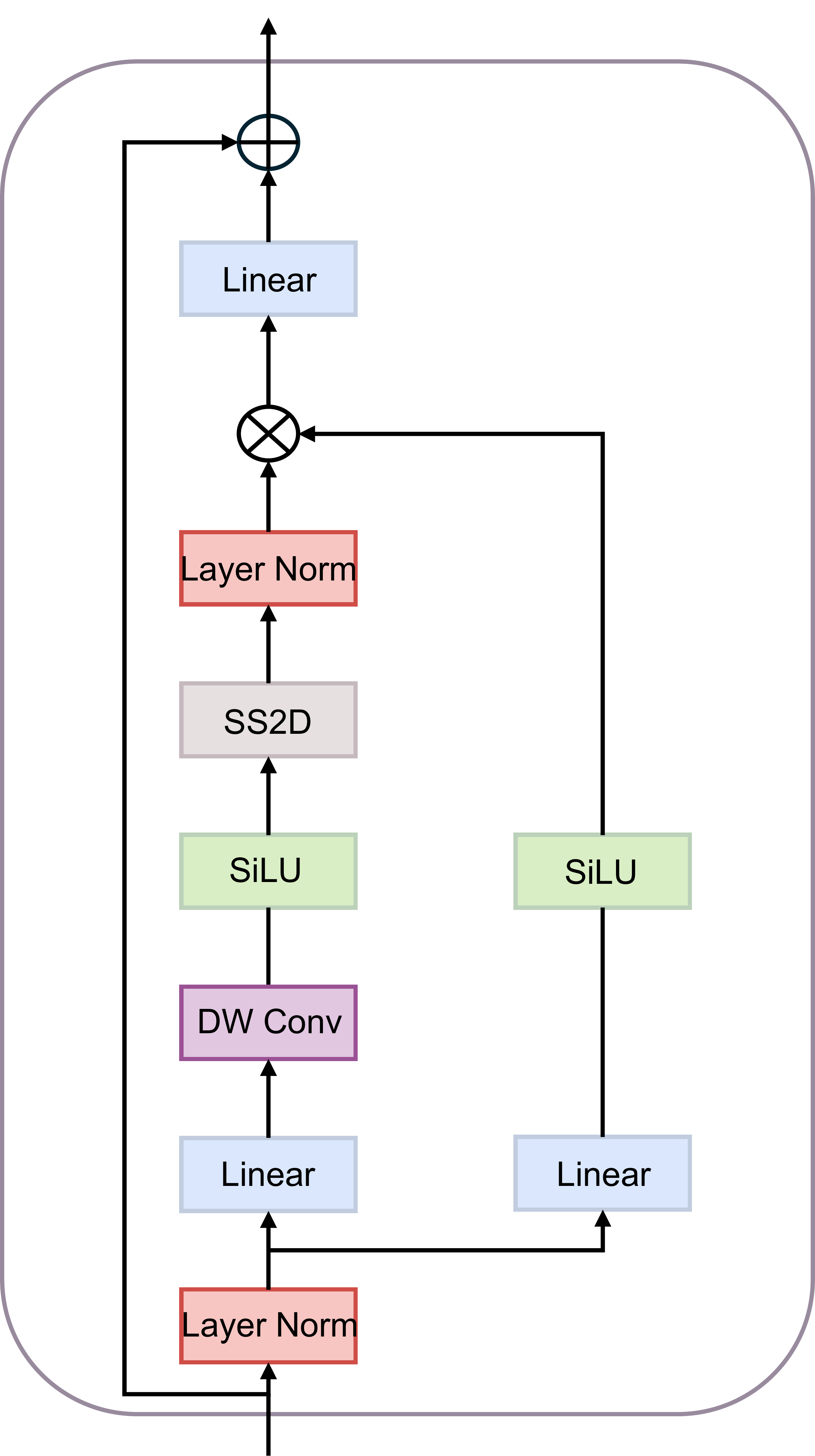} &
\includegraphics[width=0.2\textwidth]{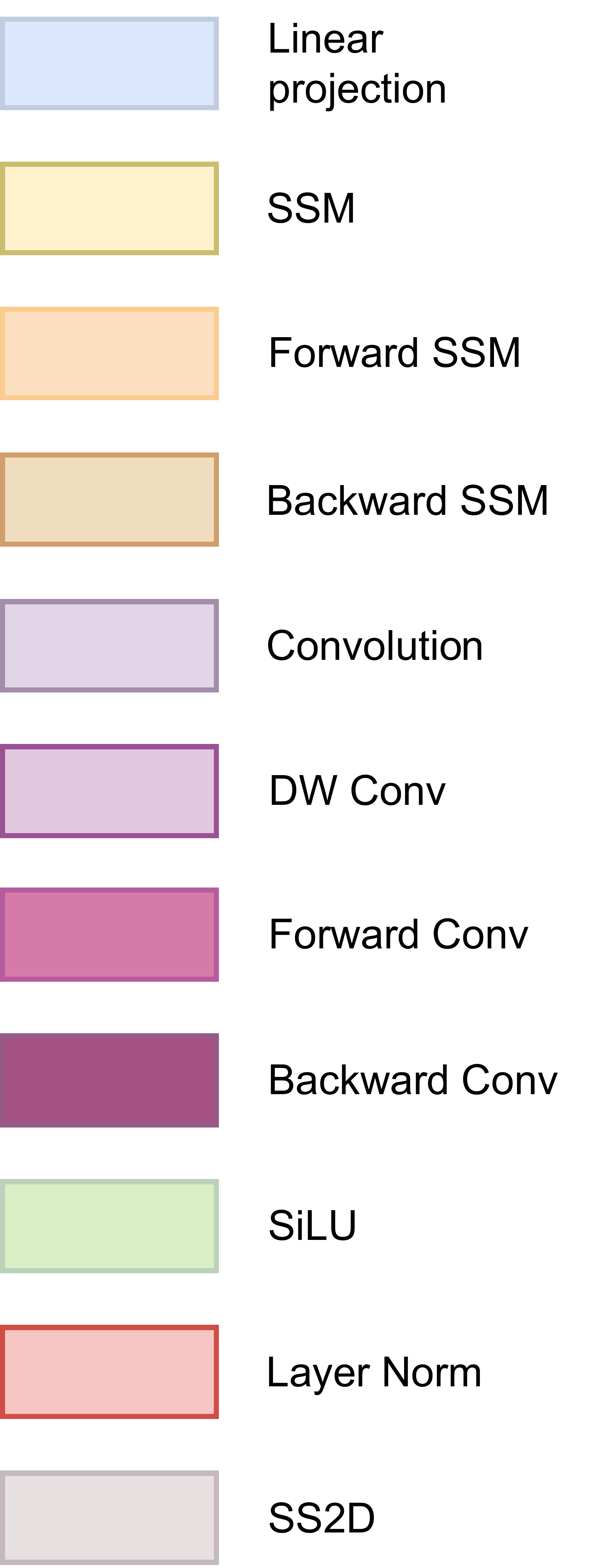} \\
(a) ViM block & (b) VSS block &
\end{tabular}
\caption{ViM Block and VSS Block.} \label{fig:vim-vss}
\end{figure}

\paragraph{ViM.} ViM block~\cite{zhu2024vision}, sometimes also mentioned as Bidirectional Mamba block, annotates image sequences with position embeddings and condenses visual representations using bidirectional state space models. It processes input both forward and backward, employing one-dimensional convolution for each direction as shown in (a) Fig.~\ref{fig:2dscanning}. The Softplus function ensures non-negative $\Delta$. Forward and backward $\vy$ are computed via the state space model described in equations (\ref{eq:dis-A-B}) and (\ref{eq:dis-ssms}), and then combined through \emph{SiLU} gating to produce the output token sequence as (a) Fig.~\ref{fig:vim-vss}.

\paragraph{VSS.} The Visual State Space (VSS) block ~\cite{liu2024vmamba} incorporates the pivotal state space model operation. It begins by directing the input through a depth-wise convolution layer, followed by a \emph{SiLU} activation function, and then through the state space model outlined in equations (\ref{eq:dis-A-B}) and (\ref{eq:dis-ssms}) employing an approximate $\overline{\vB}$. Afterwards, the output of the state space model is subjected to layer normalization before being amalgamated with the output of other information streams as (b) Fig.~\ref{fig:vim-vss}. 
To address the encountered direction-sensitive issue, they
introduce the Cross-Scan Module (CSM) to traverse the spatial domain and convert
any non-causal visual image into order patch sequences, as shown in (b) Fig.~\ref{fig:2dscanning}. They refine the approximation of $\overline{\vB}$ using the first-order Taylor series $\overline{\vB} = (\Delta \vA)^{-1}(\exp(\Delta \vA)-\vI) \cdot \Delta \vB \thickapprox (\Delta \vA)(\Delta \vA)^{-1}\Delta \vB = \Delta \vB$. 

\subsection{Pure Mamba}

\paragraph{ViM-based.}
Inspired by the vision transformer architecture, Vision Mamba~\cite{zhu2024vision} replaces the transformer encoder with a vision mamba encoder based on ViM blocks while retaining the remainder of the process. This involves converting the two-dimensional image into flattened patches, followed by linear projection of these patches into vectors and the addition of position embeddings. A class token represents the entire patch sequence, and subsequent steps involve normalization layers and a MLP layer to derive the final predictions.

\begin{figure}[htp]
\begin{tabular}{cccc}
   \includegraphics[width=0.2\textwidth]{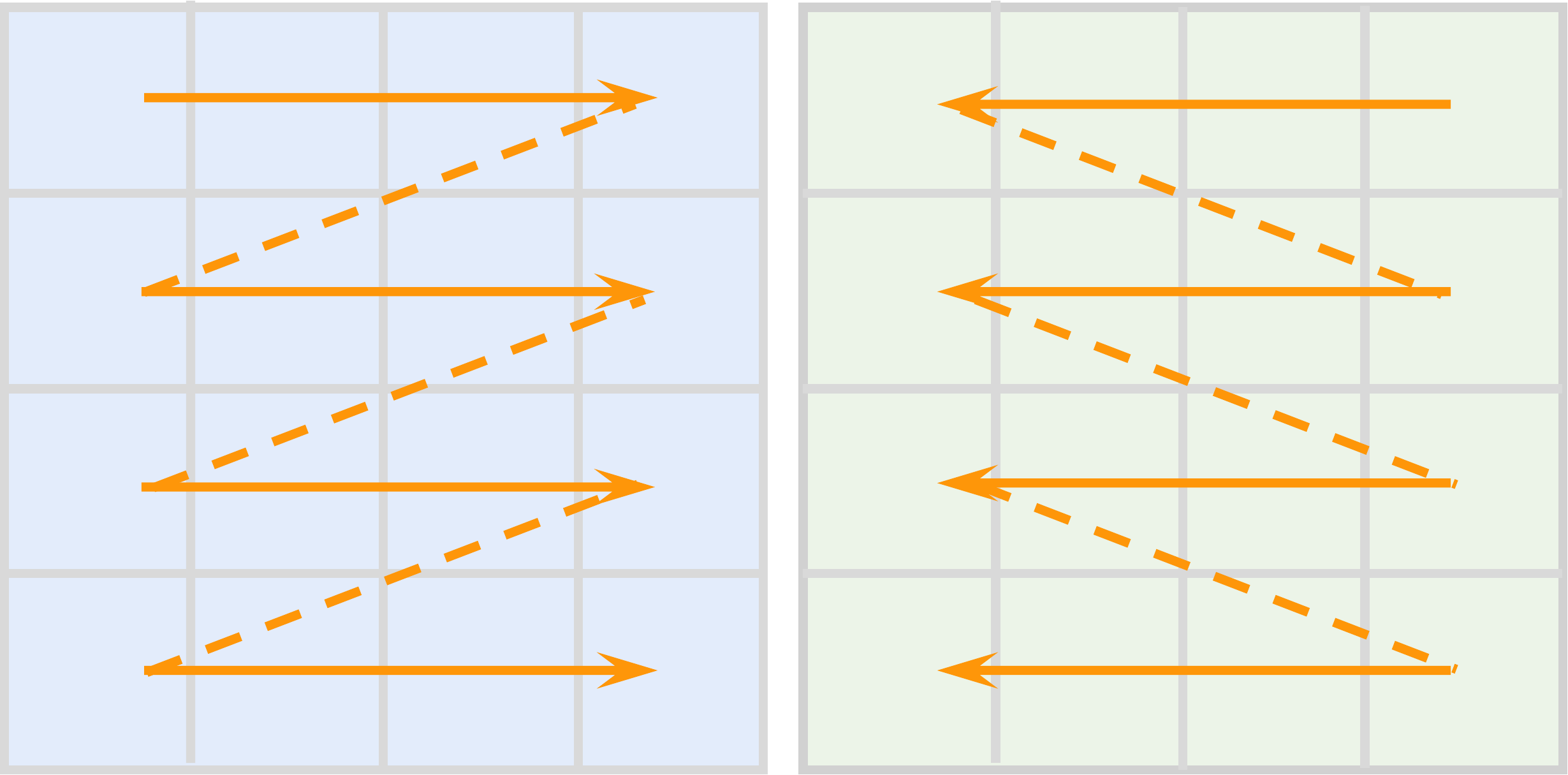}
  &  \includegraphics[width=0.2\textwidth]{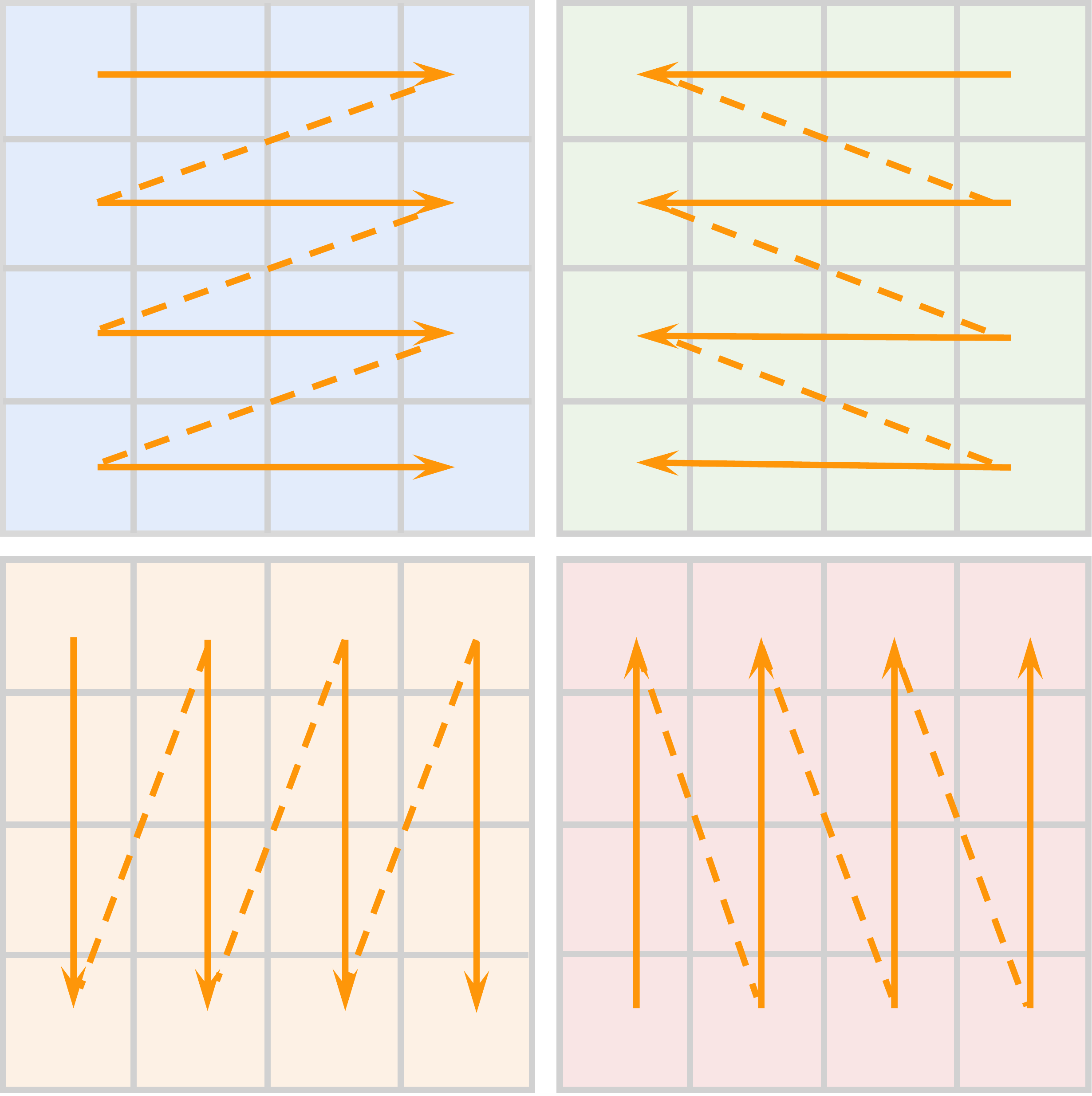}
     & \includegraphics[width=0.2\textwidth]{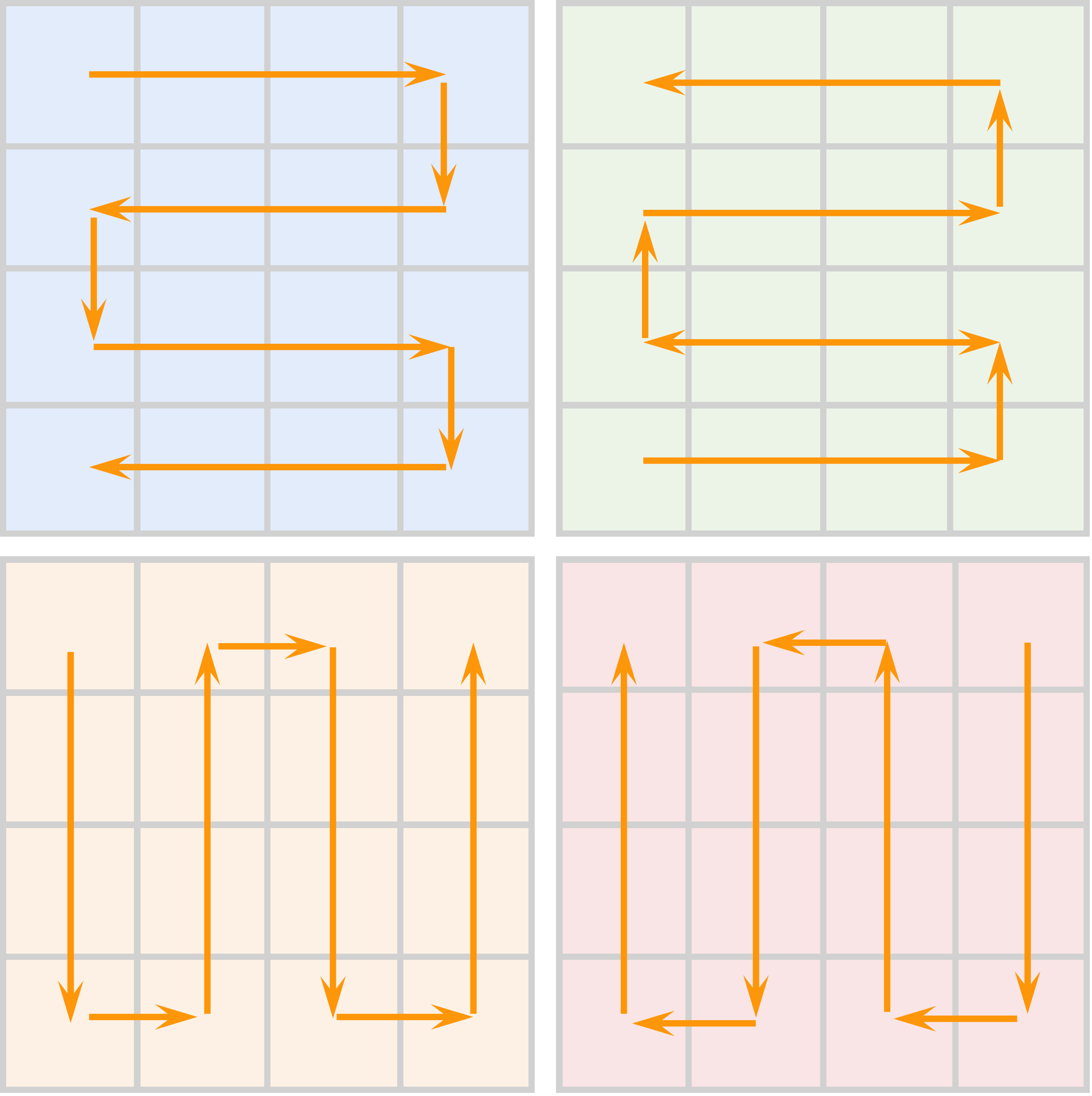}
     &
     \includegraphics[width=0.2\textwidth]{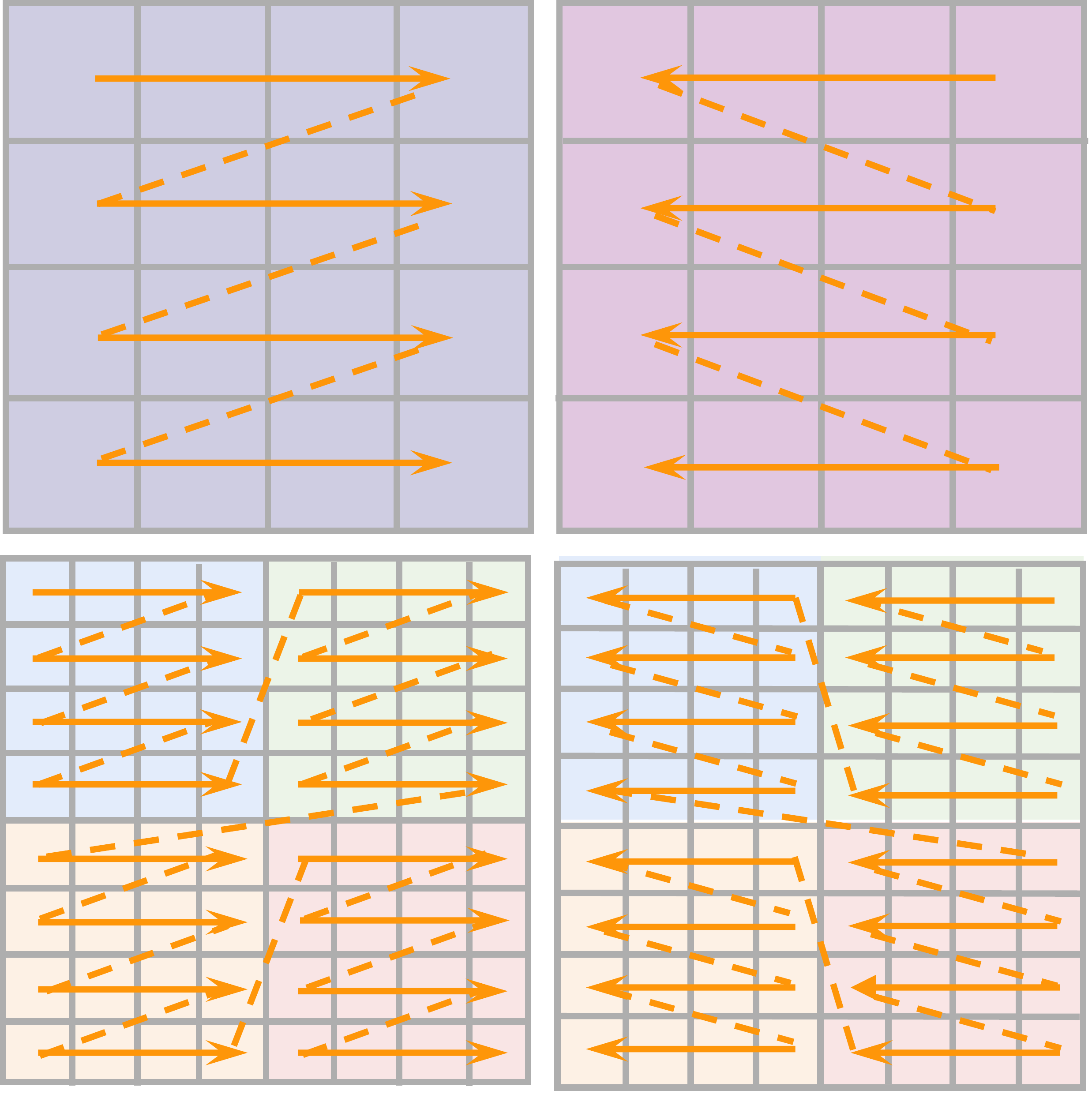}\\
     (a) BiDirectional & (b) Cross-Scan~\cite{liu2024vmamba} & (c) Continuous 2D  & (d) Local Scan~\cite{huang2024localmamba} \\
    Scan~\cite{zhu2024vision} & & ~~~Scanning~\cite{yang2024plainmamba} & \\
    ~~\\
    \mc{2}{ \includegraphics[width=0.4\textwidth]{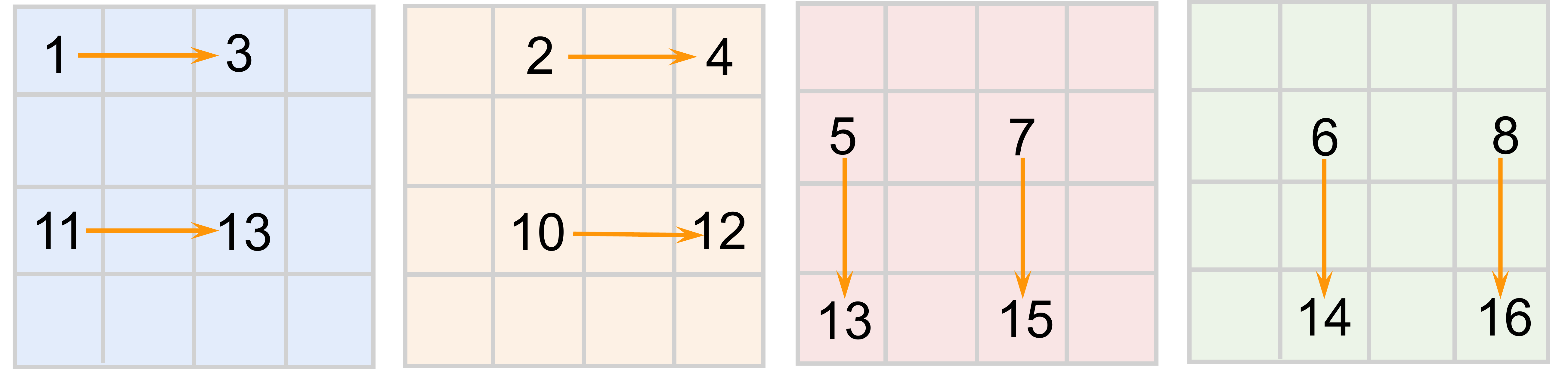}}& \mc{2}{\includegraphics[width=0.4\textwidth]{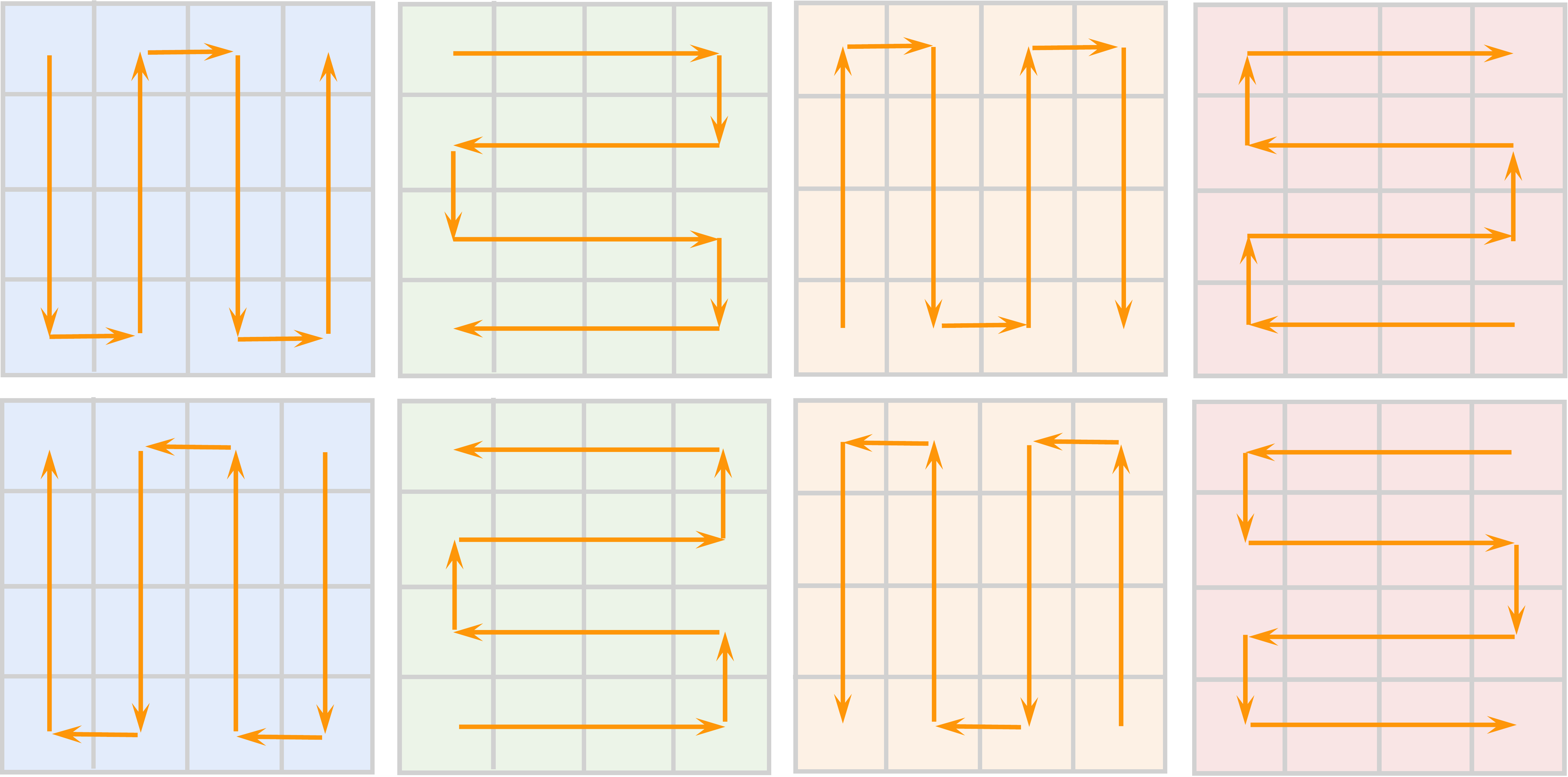}}\\
    \mc{2}{(e) Efficient 2D Scanning (ES2D)~\cite{pei2024efficientvmamba}}&\mc{2}{(f) Zigzag Scan~\cite{hu2024zigma}}\\
    ~~\\
    \mc{2}{\includegraphics[width=0.4\textwidth]{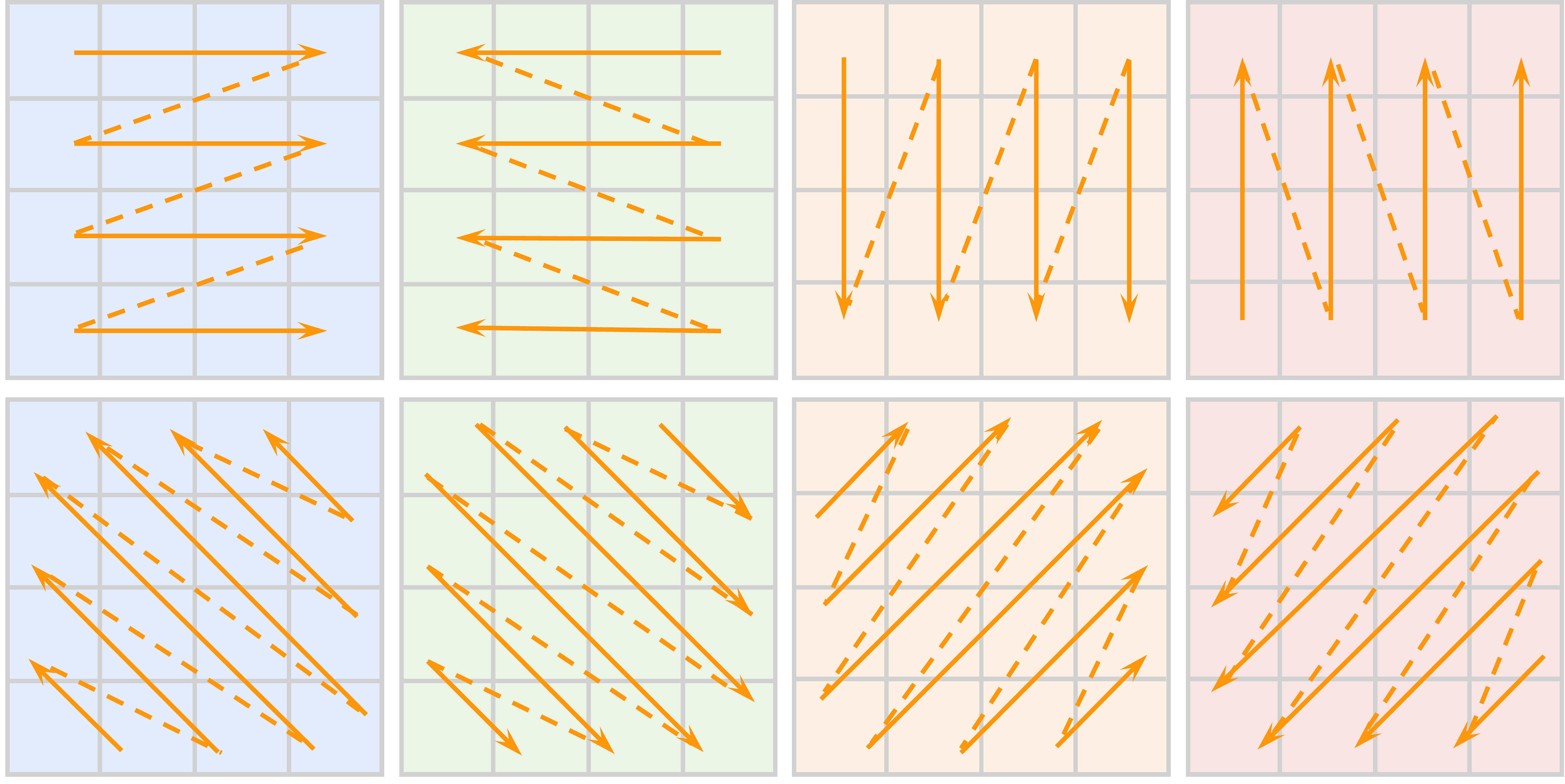}} & \mc{2}{\includegraphics[width=0.4\textwidth]{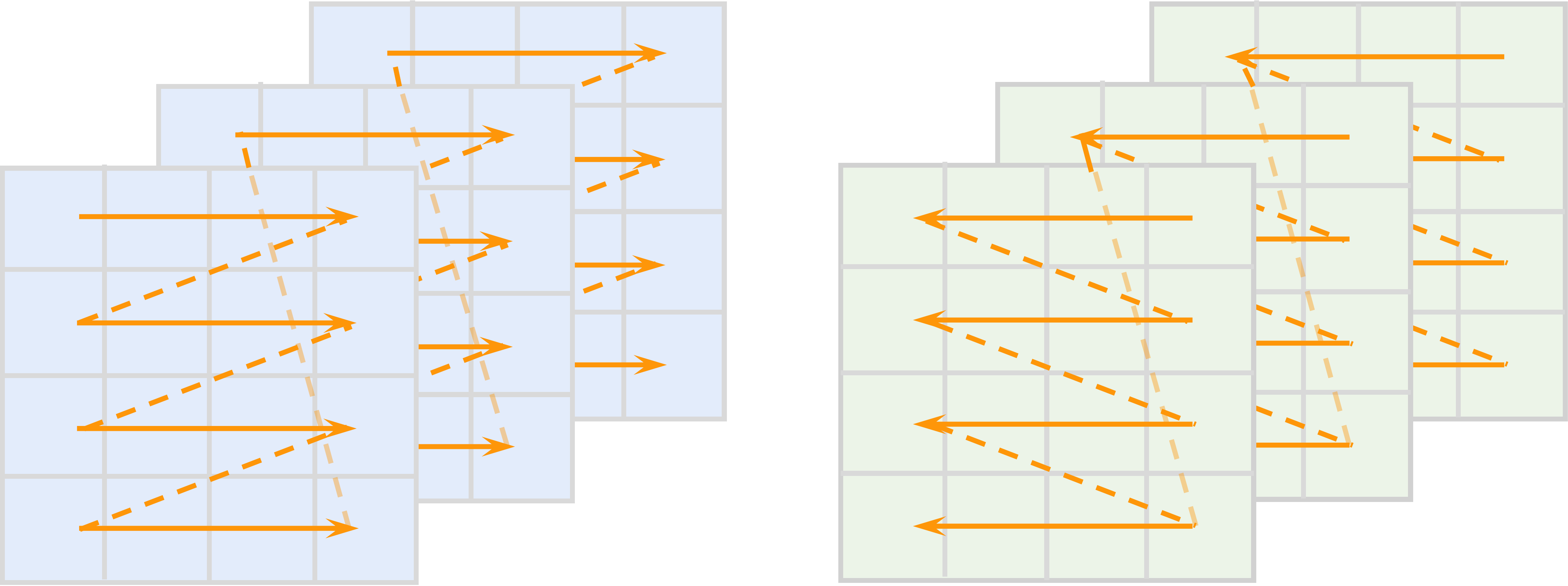}}\\
     \mc{2}{(g) Omnidirectional Selective Scan~\cite{shi2024vmambair}} & \mc{2}{(h) 3D BiDirectional Scan~\cite{li2024videomamba}}\\
     ~~\\
     \mc{2}{\includegraphics[width=0.4\textwidth]{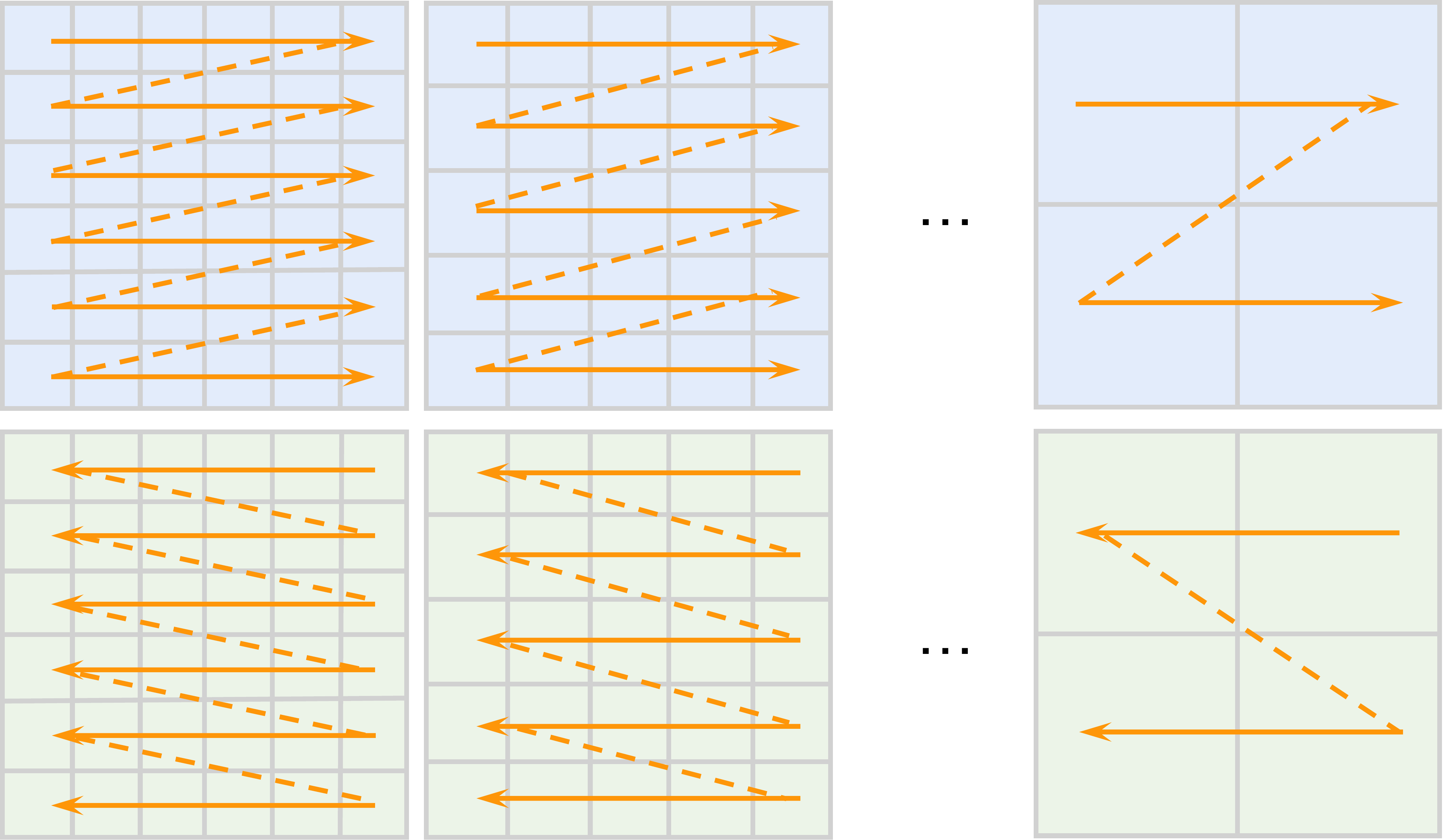}}& \mc{2}{\includegraphics[width=0.5\textwidth]{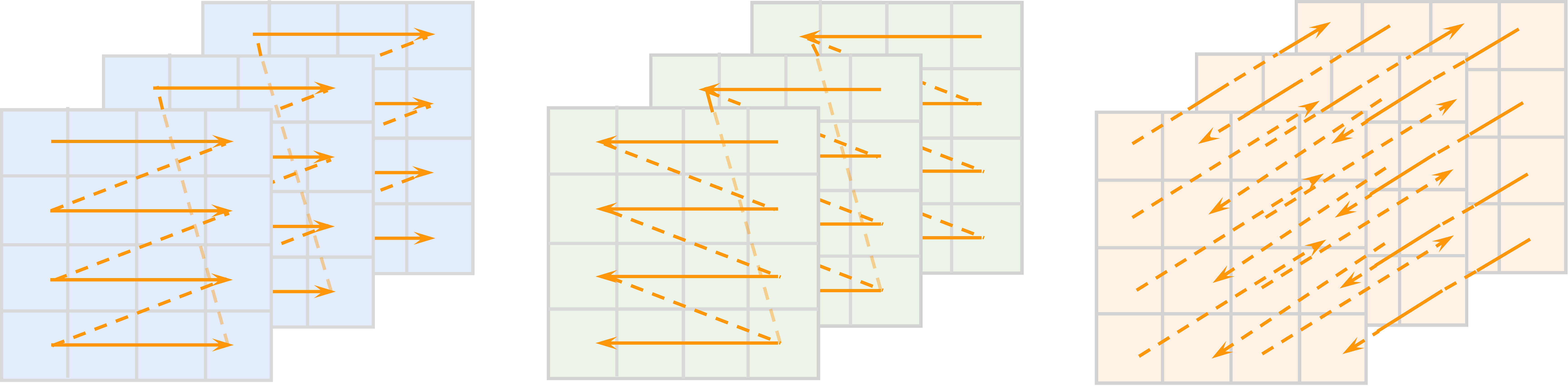}}\\
     \mc{2}{(i) Hierarchical Scan~\cite{zhang2024motion}} & \mc{2}{(j) Spatiotemporal Selective Scan~\cite{yang2024vivim}}\\
     ~~\\
     \mc{2}{\includegraphics[width=0.4\textwidth]{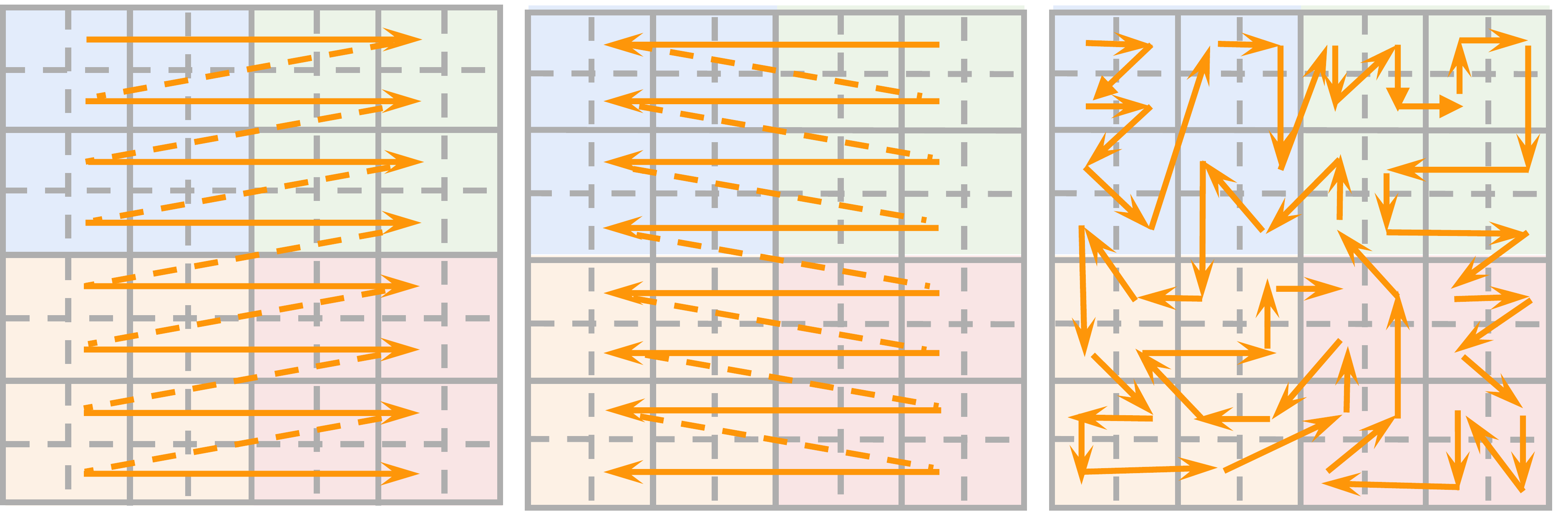}}\\
     \mc{2}{(k) Multi-Path Scan~\cite{chen2024rsmamba}}
\end{tabular}
\caption{Comparison between different 2D scanning and the selective scan orders in Vim, VMamba, PlainMamba, LocalMamba, Efficient VMamba, Zigzag, VmambaIR, VideoMamba, Motion Mamba, Vivim and RSMamba.} \label{fig:2dscanning}
\end{figure}

LocalMamba~\cite{huang2024localmamba} is built based on Vim block and it introduces a novel scanning methodology that includes localized scanning within distinct windows to capture detailed local information in conjunction with global context. Plus, LocalMamba searches scanning directions across different network layers to identify and apply the most effective scanning combinations. They propose two variants, \ie with plain and hierarchical structures.
They proposed their LocalVim Block including four scanning directions (\cf (d) Fig.~\ref{fig:2dscanning}): vim scanning and partitions tokens into distinct windows along with their flipped counterparts facilitating scanning from tail to head, state space module, and spatial and channel attention module (SCAttn).

\paragraph{VSS-based.}
VMamba~\cite{liu2024vmamba} has four stages after partitioning the input image into patches as Vision Mamba. VMamba stacks several VSS blocks on the feature map with resolution $\frac{H}{4} \times \frac{W}{4}$ as \emph{Stage 1}. In \emph{Stage 2}, before more VSS blocks involving, the feature map in \emph{Stage 1} goes through a patch merge operation for down-sampling in order to build hierarchical representations, resulting in an output resolution of $\frac{H}{8} \times \frac{W}{8}$. While \emph{Stage 3} and \emph{Stage 4} are the repetition of \emph{Stage 1} and \emph{Stage 2} with resolutions of $\frac{H}{16} \times \frac{W}{16}$ and $\frac{H}{32} \times \frac{W}{32}$.

Based on VSS block, PlainMamba block~\cite{yang2024plainmamba} enhances its capability to learn features from two-dimensional images through two main mechanisms: (i) employing a continuous 2D scanning process to improve spatial continuity, ensuring tokens in the scanning sequence are adjacent, as illustrated in (c) Fig.~\ref{fig:2dscanning}, and (ii) incorporating direction-aware updating to enable the model to discern spatial relations among tokens by encoding directional information. 
PlainMamba improves the spatial discontinuity when moving to a new row/column in the 2D scanning mechanisms of Vim and VMamba by continuing the scanning with a reversed direction until it reaches the final vision token of the image. Furthermore, PlainMamba eliminates the need for special tokens.

Within lightweight model designs, EfficientVMamba~\cite{pei2024efficientvmamba} improves the capabilities of VMamba with an atrous-based selective scan approach, \ie Efficient 2D Scanning (ES2D). Instead of scanning all patches from various directions and increasing the total number of patches, ES2D adopts a strategy of scanning forward vertically and horizontally while skipping patches and maintaining the number of patches unchanged, as shown in (e) Fig.~\ref{fig:2dscanning}. Their Efficient Visual State Space (EVSS) block comprises a convolutional branch for local features, uses ES2D as the SSM branch for global features, and all branches end through a squeeze-excitation block. 
They employ EVSS blocks for both Stage 1 and Stage 2, while opting for Inverted Residual blocks in Stage 3 and Stage 4 to enhance the capture of global representations.

\paragraph{Visual data as multi-dimensional data.}
As a part of multi-dimensional data, existing models for multi-dimensional data also work for visual-related tasks but often lack the capacity to facilitate inter- and intra-dimension communication or data-independent. The MambaMixer block~\cite{behrouz2024mambamixer} introduces a dual selection mechanism spanning across tokens and channels. It then links the sequential selective mixers through a weighted averaging mechanism, enabling layers to directly access input and output from various layers.
Mamba-ND~\cite{li2024mamba} expands the application of the SSM to higher dimensions by alternating sequence wandering across layers. Utilizing a similar scanning strategy as VMamba in the 2D scenario, it extends this approach to 3D. Additionally, they advocate for the use of multi-head SSM as an analogue to multi-head attention.
In response to the inefficiencies and performance challenges encountered by traditional transformers in image and time series processing, a new architecture named Simplified Mamba-based Architecture, \emph{SiMBA}~\cite{patro2024simba} is proposed to incorporate the Mamba block for sequence modeling and Einstein FFT (EinFFT) for channel modeling, with the goal of enhancing the stability and efficiency of the model in handling image and time series tasks. The Mamba block proves effective at processing long sequence data, while EinFFT represents a novel channel modeling technique. Experimental results demonstrate that SiMBA surpasses existing State Space Models and transformers across multiple benchmark tests.

\paragraph{Summary of 2D scanning mechanisms.}
Scan is a key component for Mamba, when it comes to multi-dimensional inputs, the scanning mechanisms matter. We summarize the existing 2D scanning mechanisms in Fig.~\ref{fig:2dscanning}. 
In particular, the direction-aware updating employs a set of learnable parameters $\{\Theta_k\}$ to representing the four cardinal directions plus a special begin direction for the initial token, and reformulate (\ref{eq:dis-ssms}) as
\begin{align}
\begin{split}
    h'_k(t) = &~ \overline{\vA}_t h_k(t) + ( \overline{\vB}_t + \overline{\Theta}_{k,t}) x(t), \\
    y'(t) = &~ \sum_{k=1}^4 \vC_t h'_k(t),\\
    y(t) =&~ y'(t) \odot z(t).     
\end{split}
\label{eq:plainmamba}
\end{align}
Expanding on the fundamental structure of Mamba and (\ref{eq:plainmamba}), we can devise the additional scanning mechanisms depicted in Fig.~\ref{fig:2dscanning}.

As an important element of Mamba, scanning mechanisms not only help in efficiency but also provide information in the scenario of visual-related tasks. We summarize the usage of different scanning mechanisms in existing works as Table.~\ref{tab:scanning}. Cross-Scan~\cite{liu2024vmamba} and BiDirectional Scan~\cite{zhu2024vision} stand out as the most widely adopted scanning mechanisms. However, various other scanning mechanisms serve specific purposes. For example, 3D BiDirectional Scan~\cite{li2024videomamba} and Spatiotemporal Selective Scan~\cite{yang2024vivim} are tailored for video inputs. Local Scan~\cite{huang2024localmamba} focuses on gathering local information, while ES2D~\cite{pei2024efficientvmamba} prioritizes efficiency.

\begin{table}[H]
\centering
\caption{Summary of the Scanning mechanisms used in visual Mamba.} 
\label{tab:scanning} 
\begin{tabular}{l| r}
\toprule
\textbf{Scanning Mechanisms} & \textbf{Method}  \\
\midrule
\multirow{5}{*}{BiDirectional Scan~\cite{zhu2024vision}} & Vision Mamba~\cite{zhu2024vision},Motion Mamba~\cite{zhang2024motion}\\ &HARMamba~\cite{li2024harmamba},MMA~\cite{cheng2024activating},VL-Mamba\cite{qiao2024vl}\\ &Video Mamba Suite~\cite{chen2024video},Point Mamba~\cite{liu2024point}\\ &LMa-UNet~\cite{wang2024large}\\ &Motion-Guided Dual-Camera Tracker~\cite{zhang2024motion2}\\\midrule
\multirow{9}{*}{Cross-Scan~\cite{liu2024vmamba}} & VMamba~\cite{liu2024vmamba},VL-Mamba\cite{qiao2024vl},VMRNN~\cite{tang2024vmrnn}\\ &RES-VMAMBA~\cite{chen2024res},Sigma~\cite{wan2024sigma},ReMamber~\cite{yang2024remamber}\\ &Mamba-UNet~\cite{wang2024mamba},Semi-Mamba-UNet~\cite{ma2024semi}\\ &VMambaMorph~\cite{wang2024vmambamorph},ChangeMamba~\cite{chen2024changemamba}\\ &H-vmunet~\cite{wu2024h},MambaMIR~\cite{huang2024mambamir},MambaIR~\cite{guo2024mambair}\\ &Serpent~\cite{shahab2024serpent},Mamba-HUNet~\cite{shahriar2024integrating},TM-UNet~\cite{tang2024rotate}\\ &Swin-UMamba~\cite{liu2024swin},UltraLight VM-UNet~\cite{wu2024ultralight}\\ &VM-UNet~\cite{ruan2024vm},VM-UNET-V2 ~\cite{zhang2024vm}\\ &MedMamba~\cite{yue2024medmamba},MIM-ISTD~\cite{chen2024mim},RS3Mamba~\cite{ma2024rs3mamba}\\\midrule
Continuous 2D Scanning~\cite{yang2024plainmamba} & PlainMamba~\cite{yang2024plainmamba}\\\midrule
Local Scan~\cite{huang2024localmamba} &LocalMamba~\cite{huang2024localmamba},FreqMamba~\cite{zhen2024freqmamba}\\\midrule
Efficient 2D Scanning (ES2D)~\cite{pei2024efficientvmamba}& EfficientVMamba~\cite{pei2024efficientvmamba}\\\midrule
Zigzag Scan~\cite{hu2024zigma}& ZigMa~\cite{hu2024zigma}\\\midrule
Omnidirectional Selective Scan~\cite{shi2024vmambair} & VmambaIR~\cite{shi2024vmambair},RS-Mamba~\cite{zhao2024rs}\\\midrule
3D BiDirectional Scan~\cite{li2024videomamba} & VideoMamba~\cite{li2024videomamba}\\\midrule
Hierarchical Scan~\cite{zhang2024motion} & Motion Mamba~\cite{zhang2024motion}\\\midrule
Spatiotemporal Selective Scan~\cite{yang2024vivim} & Vivim~\cite{yang2024vivim}\\\midrule
Multi-Path Scan~\cite{chen2024rsmamba} & RSMamba~\cite{chen2024rsmamba}\\
\bottomrule
\end{tabular}
\end{table}

\subsection{Mamba with Other Architectures}

Mamba, being a novel component compared to convolution, recurrence, and attention, offers opportunities for synergistic combinations with other architectures that are still relatively underexplored. In this section, we examine existing exploratory findings on such combinations. 
% We consolidate these combinations in Table~\ref{tab:mamba-with-cnnrnnatention}.

\paragraph{Mamba with Convolution.}
To combine Mamba with convolution, Mamba introduces the ability to obtain local information, which is essential for tasks related to medical images or segmentation tasks.
RES-VMAMBA~\cite{chen2024res} pioneers incorporating a residual learning framework within the VMamba model to simultaneously leverage global and local state features inherent in the original VMamba architectural design. This architecture commences with a stem module responsible for processing the input image, followed by a series of VSS Blocks organized sequentially across four distinct stages. Diverging from the original VMamba framework, the Res-VMamba architecture adopts the VMamba structure as its backbone and directly integrates raw data into the feature map. They refer to this integration as the global-residual mechanism to distinguish it from the residual structure in the VSS block. This integration aims to facilitate the sharing of global image features alongside the information processed through the VSS blocks. This design seeks to harness the localized details captured by individual VSS blocks and the overarching global features inherent in the unprocessed input, thereby enhancing the model's representational capacity and improving its performance on tasks requiring a comprehensive understanding of visual data.

\paragraph{Mamba with Recurrence.}
To harness the long-sequence modeling capabilities of Mamba blocks and the spatiotemporal representation prowess of LSTM, the \emph{VMRNN}\cite{tang2024vmrnn} Cell eliminates all weights and biases in ConvLSTM\cite{shi2015convolutional} and employs VSS blocks to learn spatial dependencies vertically. In the VMRNN Cell, long-term and short-term temporal dependencies are captured by updating the information on cell states and hidden states from a horizontal perspective. Building upon the VMRNN Cell, two variants are proposed: VMRNN-B and VMRNN-D. VMRNN-B primarily relies on stacking VMRNN layers, while VMRNN-D incorporates more VMRNN Cells and introduces Patch Merging and Patch Expanding layers.
The Patch Merging layer serves for downsampling, effectively reducing the spatial dimensions of the data, which aids in decreasing computational complexity and capturing more abstract, global features. Conversely, the patch-expanding layer is utilized for upsampling, increasing the spatial dimensions to restore detail and enable precise localization of features in the reconstruction phase. Ultimately, the reconstruction layer takes the hidden state from the VMRNN layer and scales it back to the input size, generating the predicted frame for the next time step. Integrating downsampling and upsampling processes offers significant advantages in our predictive architecture. Downsampling simplifies the input representation, enabling the model to process higher-level features with reduced computational overhead. This is particularly advantageous for more abstractly understanding complex patterns and relationships within the data.

\paragraph{Mamba with Attention.}
The \emph{SSM-ViT} block \cite{zubic2024state} is introduced for effective event-based information processing. It comprises three main components: a self-attention block (\emph{Block-SA}), a dilated attention block (\emph{Grid-SA}), and an SSM block. Block-SA focuses on immediate spatial relations and provides a detailed representation of nearby features. Grid-SA offers a global perspective, capturing comprehensive spatial relations and overall input structure. The SSM block ensures temporal consistency and smooth information transition between consecutive time steps. By integrating SSM with self-attention, the \emph{SSM-ViT} block enables faster training and parameter timescale adjustment for temporal aggregation.

The Meet More Areas (\emph{MMA}) block introduced in \cite{cheng2024activating} adopts a MetaFormer-style architecture, comprising two Layer Normalization layers, a token mixer (consisting of a channel attention mechanism and a ViM block in parallel), and an MLP block for deep feature extraction. There are two main reasons for this choice: Firstly, models adopting MetaFormer-style architectures have shown promising results, indicating the potential for achieving favorable outcomes. Secondly, to fully leverage and utilize the global information extracted by the ViM block, the channel attention mechanism is incorporated to activate more pixels, as global details play a role in determining the channel attention weights. Additionally, it's reasonable to suggest that employing a convolution-based module can enhance the visual representation obtained by the ViM block and streamline the training process, similar to the benefits observed with transformers. For restoration, Residual State Space Blocks (RSSBs)~\cite{guo2024mambair} block add VSS block in front of the channel attention block, which enables VSS to focus on learning diverse channel representation after which the critical channels are selected by subsequent channel attention, thus avoiding channel redundancy.

% \paragraph{U-Net.}
% UVM-Net~\cite{zheng2024u} combines convolutional layers and SSM by employing an encoder-decoder network architecture while ViM blocks as the key component rolling the feature map over the channel domain. Unlike U-Mamba ~\cite{zheng2024u} and Mamba-UNet ~\cite{wang2024mamba}, the ViM block establishes long-range dependencies on another dimension of the feature map (the non-channel domain).

% Weak-Mamba-UNet~\cite{wang2024weak} combines CNN-based UNet Vit-based SwinUNet and Mamba-based MambaUNet.

%% file: tex/ApplicationTask.tex
\section{Visual Mamba in Application Fields}

Mamba-based modules elevate the efficiency of processing sequential data, adeptly capturing long-range dependencies and seamlessly integrating into existing systems. In medical visual tasks and remote sensing images, where inputs often entail high-resolution data, Mamba emerges as a pivotal tool in augmenting various visual tasks, particularly those pertinent to medical applications.

In this section, we begin by highlighting the contributions of Mamba-based modules in enhancing general visual-related tasks. Subsequently, we delve into their specific impact on medical visual tasks and remote-sensing images.

\input{tex/GeneralTask}

\subsection{Medical Visual Mamba}

Transformers~\cite{vaswani2017attention} have profoundly influenced the field of medical imaging with their ability to master complex data representations. They have led to notable advancements across various imaging modalities, including Radiography~\cite{seeram2019digital}, Endoscopy~\cite{lui2020overview}, Computed Tomography (CT)~\cite{withers2021x}, Ultrasound Images~\cite{christensen2020super}, and Magnetic Resonance Imaging (MRI)~\cite{tiwari2020brain}. However, because most medical images are high-resolution and detailed, transformer models typically require considerable computational resources, which scale quadratically with image resolution.

The medical imaging field has recently experienced a surge in the development of Mamba-based methodologies, particularly following the introduction of VMamba. This section provides detailed examples of these design choices, further dividing them into 2D and 3D-based approaches based on the input type, as displayed in Table~\ref{medicalcv}.

\begin{figure}[H]
\includegraphics[width=\textwidth]{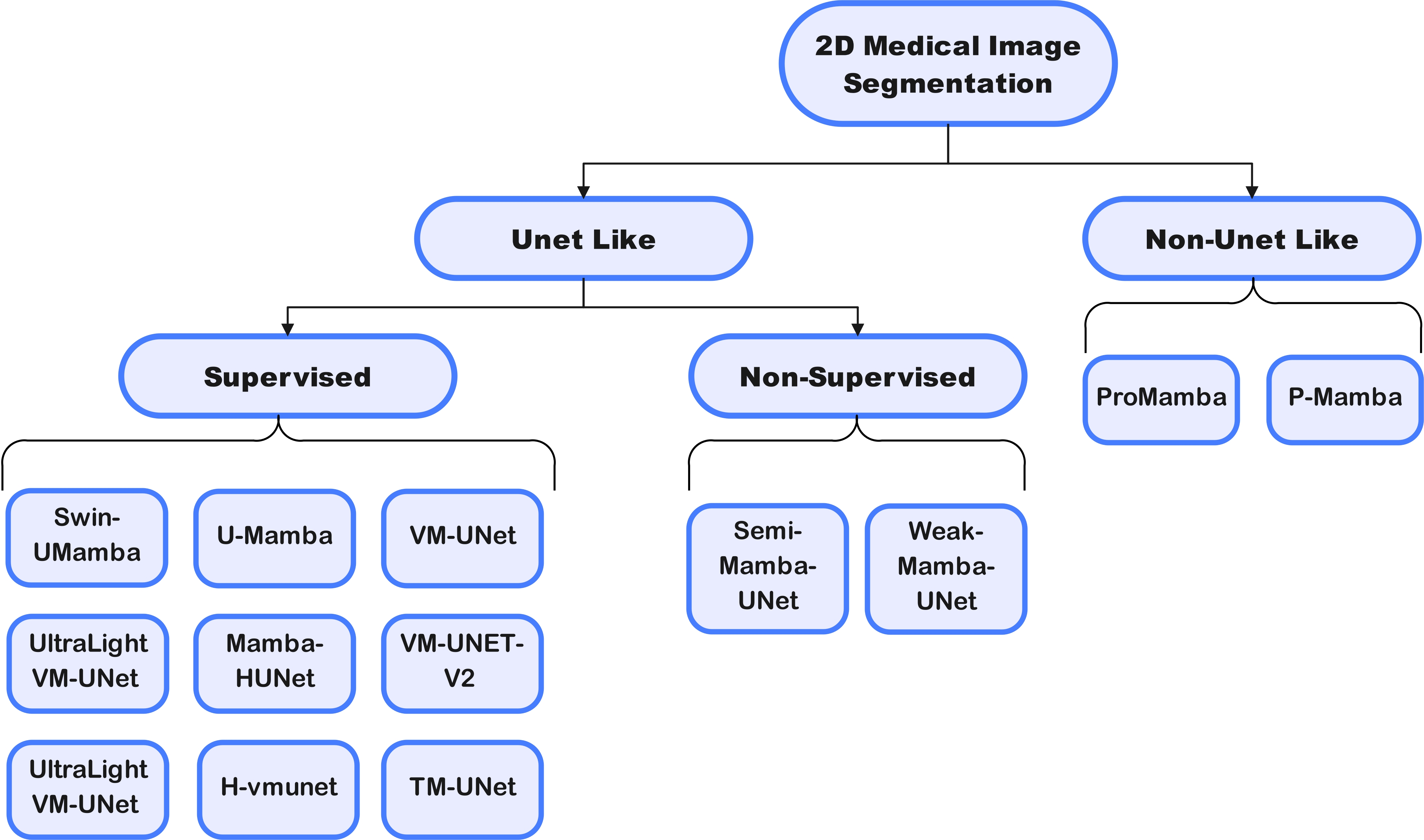}
\caption{An overview of Mamba models used for segmentation task in 2D medical images.} \label{fig:2D segmentation}
\end{figure}

\begin{table}[]
\centering
\begin{threeparttable}
\caption{Representative works of medical visual mamba} % 加入表格标题
\label{medicalcv} % 用于引用表格
% \begin{adjustwidth}{-\extralength}{0cm}
\begin{adjustwidth}{-2cm}{0cm} % Adjust the left and right margins
\begin{tabular}{c|c|c|c|c}
\toprule
\textbf{Category} & \textbf{Sub-category} & \textbf{Method} & \textbf{Efficiency} & \textbf{Code} \\
\midrule 
\multirow{34}{*}{2D}  & \multirow{24}{*}{Segmentation}  & Mamba-UNet ~\cite{wang2024mamba} & - & \Checkmark \\\cline{3-5}
  &   & H-vmunet ~\cite{wu2024h} & 
  \multicolumn{1}{c|}{\begin{tabular}[c]{@{}c@{}}Memory 0.676\\ Params 8.97\end{tabular}}
  & \Checkmark \\\cline{3-5}
 &   & Mamba-HUNet ~\cite{shahriar2024integrating} & - & \XSolid \\\cline{3-5}
  &   & P-Mamba ~\cite{ye2024p} &  \multicolumn{1}{c|}{\begin{tabular}[c]{@{}c@{}}Inference speed 23.49\\Memory 12.22\\ Params 183.37\\  FLOPs 71.81×$10^{9}$ \end{tabular}}
  & \XSolid \\\cline{3-5}
  &   & ProMamba ~\cite{xie2024promamba} & Params 102 & \XSolid \\\cline{3-5}
  &   & TM-UNet ~\cite{tang2024rotate} & 
\multicolumn{1}{c|}{\begin{tabular}[c]{@{}c@{}}Params 14.86\\ Total Params 8.41 \\ FLOPs 3.42\end{tabular}}
  & \XSolid \\\cline{3-5}
  &   & Semi-Mamba-UNet ~\cite{ma2024semi} & - & \Checkmark \\\cline{3-5}
  &   & Swin-UMamba ~\cite{liu2024swin} & \multicolumn{1}{c|}{\begin{tabular}[c]{@{}c@{}}Params 28\\ FLOPs 18.9\end{tabular}}
  & \Checkmark \\\cline{3-5}
  &   & UltraLight VM-UNet ~\cite{wu2024ultralight} &\multicolumn{1}{c|}{\begin{tabular}[c]{@{}c@{}}Params 0.049\\ GFLOPs 0.060\end{tabular}}
  & \Checkmark \\\cline{3-5}
  &   & U-Mamba ~\cite{ma2024u} & - & \Checkmark \\\cline{3-5}
  &   & VM-UNet  ~\cite{ruan2024vm} &  
\multicolumn{1}{c|}{\begin{tabular}[c]{@{}c@{}}Params 34.62 \\ FLOPs 7.56\\ FPS 20.612\end{tabular}}
  & \Checkmark \\\cline{3-5}
  &   & VM-UNET-V2 ~\cite{zhang2024vm} & 
\multicolumn{1}{c|}{\begin{tabular}[c]{@{}c@{}}Params 17.91 \\ FLOPS 4.40\\ FPS 32.58\end{tabular}}
  & \Checkmark \\\cline{3-5}
  &   & Weak-Mamba-UNet ~\cite{wang2024weak} & - & \Checkmark \\\cmidrule{2-5} 
  & Radiation dose prediction & MD-Dose ~\cite{fu2024md} &\multicolumn{1}{c|}{\begin{tabular}[c]{@{}c@{}}Inference speed 18\\ Params 30.47\end{tabular}}  
  & \Checkmark \\\cmidrule{2-5}
  & \multirow{2}{*}{Classification}  & MedMamba ~\cite{yue2024medmamba} & - & \Checkmark \\\cmidrule{3-5}
  &  & MambaMIL ~\cite{yang2024mambamil} & - & \Checkmark  \\\cmidrule{2-5} 
  & Image reconstruction &  MambaMIR/MambaMIR-GAN ~\cite{huang2024mambamir} & - & \Checkmark \\\cmidrule{2-5}
  & Exposure correction & FDVM-Net ~\cite{zheng2024fd} & Inference speed 22.95 & \Checkmark \\\midrule
\multirow{18}{*}{3D}  & \multirow{6}{*}{Segmentation}  & LMa-UNet ~\cite{wang2024large} & - & \Checkmark \\\cline{3-5}
  &   & LightM-UNet ~\cite{liao2024lightm} & \begin{tabular}[c]{@{}c@{}}Params 1.87\\ FLOPs 457.62×$10^{9}$ \end{tabular} 
  & \Checkmark \\\cline{3-5}
  &   & SegMamba ~\cite{xing2024segmamba} & Inference speed 151 & \Checkmark \\\cline{3-5}
  &   & T-Mamba ~\cite{hao2024t} & - & \Checkmark \\\cline{3-5}
 &    & Vivim ~\cite{yang2024vivim} & FPS 35.33  & \Checkmark \\\cmidrule{2-5} 
& Classification  & CMViM~\cite{yang2024cmvim} & Params 50 & \Checkmark \\\cmidrule{2-5}
& Motion tracking  & Motion-Guided Dual-Camera Tracker ~\cite{zhang2024motion2} & - & \XSolid  \\\cmidrule{2-5}
& Backbone  & nnMamba ~\cite{gong2024nnmamba} & \multicolumn{1}{c|}{\begin{tabular}[c]{@{}c@{}}Params 15.55\\ FLOPs 141.14\end{tabular}}
  & \XSolid  \\\cmidrule{2-5}
&\multirow {5}{*}{Image registration} & VMambaMorph ~\cite{wang2024vmambamorph} &
\multicolumn{1}{c|}{\begin{tabular}[c]{@{}c@{}}Inference speed 19 \\ Memory 3.93\\ Params 9.64\end{tabular}}
&  \Checkmark 
\\\cmidrule{3-5}
  & &MambaMorph ~\cite{guo2024mambamorph} &
\multicolumn{1}{c|}{\begin{tabular}[c]{@{}c@{}}Inference speed 27 \\ Memory 7.60\\ Params 7.59\end{tabular}}
  & \Checkmark \\
\bottomrule
\end{tabular}
% \end{adjustwidth}
\begin{tablenotes}
\footnotesize
\item[1]  For the Efficiency, Inference speed is in ms, Memory is in Gb, Params is in M, and FLOPS is in G.
\end{tablenotes}
\end{adjustwidth}
\end{threeparttable}
\end{table}

\subsubsection{2D Medical Image}

Mamba has shown impressive potential in 2D medical segmentation, as displayed in Fig.~\ref{fig:2D segmentation}. Here, we discuss in detail some methods that explore using mamba to model the global structure information of 2D medical segmentation. 

Most of the innovative architectures that have been developed are based on U-Net, which has achieved remarkable success in various medical image segmentation tasks. U-Mamba ~\cite{ma2024u} is the first extension of the mamba model to the U-Net architecture for visual segmentation in biomedical imaging, addressing the challenge of long-range dependency modeling, which is captured by a hybrid CNN-SSM block. Wu et al. introduced the High-order Vision Mamba UNet (H-vmunet) ~\cite{wu2024h}, an improvement of U-Mamba, which utilizes a High-order 2D-selective-scan at each interaction order to bolster the learning of local features while minimizing the incorporation of redundant information. Shortly after their initial release, the team expanded their work by introducing the UltraLight VM-UNet ~\cite{wu2024ultralight}. This new iteration was developed by in-depth analysis of the critical factors affecting parameter efficiency within the Mamba framework. This resulted in a remarkably lightweight model with a mere 0.049M parameters and a computational efficiency of only 0.060 GFLOPs.In addition, Mamba-UNet~\cite{wang2024mamba} combines the encoder-decoder architecture of U-Net with the capabilities of Mamba and maintains spatial information at different network scales through jump connections. A Visual Mamba-based VSS block is used, which utilizes linear embedding layers and deep convolution to extract features while downsampling and upsampling are facilitated by multiple merge operations and extension layers for comprehensive feature learning.

Pyramid ViT (PVT) and Swin-Unet are both pioneering hierarchical designs that apply visual tasks and propose progressive shrinking pyramids and spatial-reduction attention. Drawing inspiration from PVT and Swin-Unet, Ruan et al. introduced VM-UNet ~\cite{ruan2024vm}, a foundational model for purely SSM-based segmentation in medical imaging. This model demonstrates the capabilities of SSM in medical image segmentation and consists of three primary components: an encoder, a decoder, and skip connections. Building on their previous work, the team proposed VM-UNET-V2 ~\cite{zhang2024vm}. The Visual State Space (VSS) block was introduced to capture a broader range of contextual information. The Semantics and Detail Infusion (SDI) mechanism was also implemented to enhance the fusion of low-level and high-level features. Mamba-HUNet ~\cite{shahriar2024integrating}, another multi-scale hierarchical upsampling network, incorporates the Mamba technique. It leverages Visual State Space blocks and patch merging layers to extract hierarchical features, ensuring the preservation of spatial information. TM-UNet ~\cite{tang2024rotate} introduces improvements to the bottleneck layer. This architecture proposes Triplet SSM as the bottleneck layer, representing the first attempt to use a pure SSM approach to integrate spatial and channel features. Existing Mamba-based models are primarily trained from scratch, overlooking potential benefits. Swin-UMamba ~\cite{liu2024swin}, a new Mamba-based model tailored explicitly for medical image segmentation tasks, leveraged the strengths of ImageNet-based pretraining.

Previous discussions have primarily focused on supervised learning methods, but other supervisory approaches have also been explored. Semi-Mamba-UNet ~\cite{ma2024semi} combines a visual Mamba-based U-Shape Encoder-Decoder with a traditional CNN-based UNet in a semi-supervised learning framework. It introduces a self-supervised pixel-level contrastive learning strategy using a pair of projectors to improve feature learning, particularly on unlabeled data. Weak-Mamba-UNet ~\cite{wang2024weak} is a novel weakly-supervised learning framework for medical image segmentation, combining CNNs, ViT, and the VMamba. It focuses on scribble-based annotations and uses a collaborative, cross-supervisory mechanism with pseudo labels for iterative network learning and refinement.

Some segmentation approaches diverge from UNet architectures. P-Mamba ~\cite{ye2024p} introduces a novel dual-branch framework for highly efficient left ventricle segmentation in pediatric echocardiograms. This model features an innovative DWT-based encoder branch equipped with Perona-Malik Diffusion (PMD) Blocks. Moreover, P-Mamba adopts vision mamba layers within its vision mamba encoder branch to bolster computational and memory efficiency. PromptMamba ~\cite{xie2024promamba} represents a groundbreaking integration of the Vision-Mamba and prompt technologies, marking a significant milestone as the first model to leverage the Mamba framework for the specific task of polyp segmentation.

In addition, mamba has expanded its research in 2D medical imaging beyond segmentation, enhancing the precision and speed of image analysis to support diagnosis and treatment planning. Classification is a vital and fundamental task in the field of medical image analysis. Yue et al. introduced Vision Mamba for this purpose, also known as MedMamba ~\cite{yue2024medmamba}. They developed a novel module named Conv-SSM, which combines the local feature extraction of convolutional layers with the long-range dependency capture of SSM, enabling efficient modeling of medical images from different modalities. Furthermore, MambaMIL ~\cite{yang2024mambamil} introduces the Sequence Reordering Mamba (SR-Mamba), a model that recognizes the order and distribution of instances in long sequences to harness the embedded valuable information effectively. Image reconstruction is pivotal in enhancing diagnostic processes, as high-quality and high-fidelity medical images are fundamental to the accuracy and efficiency of clinical decisions. Huang et al. ~\cite{huang2024mambamir} have developed MambaMIR, a model leveraging Mamba technology for the reconstruction of medical images, alongside its advanced counterpart, MambaMIR-GAN, which incorporates Generative Adversarial Networks. Moreover, Zheng et al. introduce FDVision Mamba (FDVM-Net) ~\cite{zheng2024fd}, a frequency-domain-based network that effectively corrects image exposure by reconstructing the frequency domain of endoscopic images, as recorded endoscopic images often suffer from exposure abnormalities. In specialized areas, MD-Dose ~\cite{fu2024md}, a cutting-edge diffusion model based on the Mamba architecture, was designed to predict radiation therapy dose distribution for thoracic cancer patients accurately. 

\subsubsection{3D Medical Image}
3D image analysis in medical imaging enables more accurate and comprehensive diagnoses by providing detailed views of complex anatomical structures. Gong et al. present nnMamba ~\cite{gong2024nnmamba}, an innovative architecture designed for 3D medical imaging applications, which integrates local and global relationship modeling via the MICCSS (Mamba-In-Convolution with Channel-Spatial Siamese input) module. nnMamba was tested on a comprehensive benchmark of six datasets for three crucial tasks: segmentation, classification, and landmark detection, showcasing its capability in long-range relationship modeling at channel and spatial levels.

Precise 3D segmentation outcomes can alleviate physicians' diagnostic workloads in disease management. SegMamba~\cite{xing2024segmamba}, a cutting-edge architecture, is the first method to utilize Mamba specifically for accurate 3D segmentation in medical imaging. It introduced a tri-orientated Mamba (ToM) module for modeling 3D features from three directions and a gated spatial convolution (GSC) module to enhance spatial feature representation before each ToM module. Similarly employing a U-shaped architecture, LightM-UNet~\cite{liao2024lightm} harnesses the Residual Vision Mamba Layer exclusively in a solely Mamba approach for extracting deep semantic features and modeling extensive spatial dependencies within a lightweight framework. Moreover, both LMa-UNet~\cite{wang2024large} and T-Mamba~\cite{hao2024t} built upon the foundation of SegMamba, with improvements made to the Mamba block. A notable aspect of LMa-UNet~\cite{wang2024large} is its use of large windows, which outperforms small kernel-based CNNs and small window-based Transformers in local spatial modeling, while T-Mamba~\cite{hao2024t} develops a gate selection unit to adaptively combine two features in the spatial domain with one feature in the frequency domain, marking the first instance of incorporating frequency-based features into the vision mamba framework. The issue of long-term temporal dependency in video scenarios has also been addressed by developing a generic Video Vision Mamba-based framework named Vivim~\cite{yang2024vivim}. Using the specifically engineered Temporal Mamba Block, this model effectively compresses long-term spatiotemporal data into sequences of different scales.

For image registration task, MambaMorph~\cite{guo2024mambamorph} introduces a groundbreaking multi-modality deformable registration framework that enhances medical image analysis by combining a Mamba-based registration module with an advanced feature extractor for efficient spatial correspondence and feature learning. The VMambaMorph ~\cite{li20243dmambacomplete} has further enhanced its VMamba-based block by incorporating a 2D cross-scan module, redesigned to process 3D volumetric features efficiently.

In other domains, the Contrastive Masked Vim Autoencoder (CMViM)~\cite{yang2024cmvim} tackles Alzheimer’s disease (AD) classification by incorporating Vision Mamba (ViM) into a masked autoencoder for 3D multi-modal data reconstruction. For endoscopy skill evaluation, a low-cost motion-guided dual-camera tracker~\cite{zhang2024motion} provides reliable endoscope tip feedback, and a Mamba-based motion-guided prediction head (MMH) merges visual tracking with historical motion data using a state space model.

\subsubsection{Challenge}

Here we explore some promising future research directions for vision Mamba in medical image analysis. Challenges include the need for pre-training on large datasets, enhancing the interpretability of Mamba-based medical imaging approaches, and improving robustness against adversarial attacks. Additionally, there is a need to design efficient Mamba architectures suitable for real-time medical applications and to address the challenges in deploying Mamba-based models in distributed settings.

\begin{table}[H]
\centering
\begin{threeparttable}
\caption{Representative mamba work in remote sensing image} % 加入表格标题
\label{remote} % 用于引用表格
% \begin{adjustwidth}{-\extralength}{0cm}
\begin{tabular}{c|c|c|c|c}
\toprule
\textbf{Category} & \textbf{Method} & \textbf{Highlight} &\textbf{Efficiency} & \textbf{Code} \\
\midrule
Pan-sharpening & Pan-Mamba ~\cite{he2024pan} & \begin{tabular}[c]{@{}l@{}}channel swapping Mamba; \\ cross-modal Mamba\end{tabular} & \begin{tabular}[c]{@{}c@{}}Params 0.1827\\ FLOPs 3.0088\end{tabular} &\Checkmark \\\midrule
Infrared Small Target Detection & MIM-ISTD ~\cite{chen2024mim} & Mamba-in-Mamba architecture & \begin{tabular}[c]{@{}l@{}}Params 1.16 \\ FLOPs 1.01 \\ Inference speed 30 \\ Memory 1774\end{tabular} & \Checkmark \\ \midrule
\multirow{3}{*}{\begin{tabular}[c]{@{}l@{}} Classification\end{tabular}}&   RSMamba ~\cite{chen2024rsmamba} & multi-path activation & - & \Checkmark \\\cmidrule{2-5}
 & HSIMamba ~\cite{yang2024hsimamba} & process data bidirectionally & Memory 136.53 & \Checkmark \\\midrule
Image dense prediction & RS-Mamba ~\cite{zhao2024rs} & omnidirectional selective scan & - & \Checkmark \\ \midrule
Change detection & ChangeMamba ~\cite{chen2024changemamba} & cross-scan mechanism & - & \Checkmark \\ \midrule
\multirow{3}{*}{Semantic segmentation} & RS3Mamba ~\cite{ma2024rs3mamba} & dual-branch network & \begin{tabular}[c]{@{}l@{}}FLOPs 31.65 \\ Params 43.32 \\ Memory 2332 \end{tabular} & \Checkmark \\ \cmidrule{2-5}
 & Samba ~\cite{zhu2024samba} & encoder-decoder architecture & \begin{tabular}[c]{@{}l@{}}Params 51.9 \end{tabular} & \Checkmark \\ 
\bottomrule
\end{tabular}
\begin{tablenotes}
\footnotesize
\item[1] For the Efficiency, Inference speed is in ms, Memory is in MB, Params is in M, and FLOPS is in G.
\end{tablenotes}
\end{threeparttable}
% \end{adjustwidth}
\end{table}

\subsection{Remote Sensing Image}

The progress of remote sensing technology has sparked interest in high-resolution earth observation, with the Transformer model offering an optimal solution through its attention mechanism. However, its quadratic complexity poses challenges in modeling efficiency and memory usage. The State Space Model (SSM) addresses these issues by establishing long-distance dependencies with near-linear complexity, and Mamba further enhances efficiency through hardware optimization and time-varying parameters. The representative recent work is shown in Table~\ref{remote}. 

Drawing inspiration from TNT, Chen et al. introduced a new Mamba-in-Mamba (MiM-ISTD) ~\cite{chen2024mim} architecture to enhance the efficiency of infrared small target detection. In this approach, local patches are considered "visual sentences," while Outer Mamba is utilized to extract global information. Moreover, for remote sensing image classification, RSMamba ~\cite{chen2024rsmamba} features a dynamic multi-path activation mechanism to improve Mamba's capability in handling non-causal data. RS-Mamba ~\cite{zhao2024rs} is adept at handling very-high-resolution (VHR) remote sensing images for dense prediction tasks, utilizing an omnidirectional selective scan module to model images from various angles comprehensively. Classification of hyperspectral images is difficult in remote sensing research due to their complex, high-dimensional data.HSIMamba~\cite{yang2024hsimamba} is designed with a module dedicated to spatial analysis, which includes multiple spectral bands and three-dimensional spatial structures to take advantage of the rich multidimensional nature of the hyperspectral data and to enhance the feature representation capability using linear transformations and activation functions. In addition, HSIMamba employs a bi-directional processing mechanism that improves the model's ability to represent and utilize spectral information through forward and backward spectral dependency capture. Pan-Mamba ~\cite{he2024pan} also provides an innovative network in the pansharpening domain and customizes two crucial components, channel swapping Mamba and cross-modal Mamba, both carefully crafted for efficient cross-modal information exchange and fusion.ChangeMamba~\cite{chen2024changemamba} explores for the first time the potential of the Mamba architecture for remote sensing change detection (CD) tasks. The MambaBCD, MambaSCD, and MambaBDA network frameworks are designed for Binary Change Detection (BCD), Semantic Change Detection (SCD), and Building Damage Assessment (BDA) tasks, and three spatio-temporal relationship modeling mechanisms are proposed to learn spatio-temporal features thoroughly.ChangeMamba utilizes selective state-space modeling to capture long-range dependent features and maintains linear computational complexity while providing Visual Mamba architecture to learn global spatial context information. Semantic segmentation of remotely sensed images is a fundamental task in geoscientific research.RS3Mamba~\cite{ma2024rs3mamba} is a novel two-branch network for semantic segmentation of remotely sensed images. The network incorporates visual state space (VSS) models, especially the Mamba architecture, to improve long-range relational modeling capabilities. In addition, a Co-Completion Module (CCM) is proposed for feature fusion. Experimental results show that RS3Mamba has significant advantages over CNN and transformer-based approaches.Samba~\cite{zhu2024samba} is a novel semantic segmentation framework specialized for high-resolution remote sensing images with encoder-decoder architecture. Samba blocks act as encoders to extract multilevel semantic information, and Mamba blocks utilize state space models to capture global semantic information with linear computational complexity.

%% file: tex/GeneralTask.tex
\begin{table}[]
\centering
\begin{threeparttable}
\caption{Representative works of general visual mamba} 
\label{tab:generalcv} 
\renewcommand{\arraystretch}{0.8} %调整表格间距
% \tiny %调整表格字号
\scriptsize
% \footnotesize %
\begin{tabular}{c|c|c|c|c}
\toprule
\textbf{Category} & \textbf{Sub-category} & \textbf{Method} & \textbf{Efficiency} & \textbf{Code} \\
\midrule
\multirow{20}{*}{Backbone} & \multirow{18}{*}{Visual Mamba} &Vision Mamba ~\cite{zhu2024vision} & \begin{tabular}[c]{@{}l@{}} Params Vim-Ti: 7, Vim-S: 26 \end{tabular} &\Checkmark \\\cmidrule{3-5}
  &   & VMamba ~\cite{liu2024vmamba} & \begin{tabular}[c]{@{}l@{}} FLOPs Tiny: 4.5, Small: 9.1, Base: 15.2 \end{tabular} & \Checkmark \\\cmidrule{3-5}
 &      &  PlainMamba ~\cite{yang2024plainmamba} & \begin{tabular}[c]{@{}l@{}} FLOPs \\ PlainMamba-L1: 3.0 \\ PlainMamba-L2: 8.1 \\ PlainMamba-L3: 14.4 \end{tabular} &\Checkmark \\\cmidrule{3-5}
 &    &LocalMamba ~\cite{huang2024localmamba}  & \begin{tabular}[c]{@{}l@{}} FLOPs \\ LocalVMamba-T: 5.7 \\ LocalVMamba-S: 11.4 \end{tabular} &\Checkmark \\\cmidrule{3-5}
 &   & Mamba-ND ~\cite{li2024mamba} & \begin{tabular}[c]{@{}l@{}} Params Mamba-2D: 24, Mamba-3D: 36 \end{tabular}&\Checkmark \\\cmidrule{3-5}
& & SiMBA ~\cite{patro2024simba}    & - &\Checkmark \\\cmidrule{3-5}
 & &  RES-VMAMBA ~\cite{chen2024res}& - &\Checkmark \\\cmidrule{2-5}
  & \multirow{3}{*}{Efficient Mamba} & EfficientVMamba~\cite{pei2024efficientvmamba}& - &\Checkmark \\\cmidrule{3-5}
   &  & MambaMixer~\cite{behrouz2024mambamixer}& - &\Checkmark \\
 \midrule
\multirow{24}{*}{High/Mid-level vision} 
& Object detection&  SSM-ViT~\cite{zubic2024state}& Params 17.5 &\XSolid  \\\cmidrule{2-5}
& \multirow{3}{*}{Segmentation}&    ReMamber ~\cite{yang2024remamber}& -
 &\XSolid  \\\cmidrule{3-5}
&  & Sigma ~\cite{wan2024sigma}& - &\Checkmark \\
\cmidrule{2-5}
& Video classification&   ViS4mer ~\cite{islam2022long}& Memory 5273.6 &\Checkmark  \\\cmidrule{2-5}
 & \multirow{6}{*}{Video understanding}  & Video Mamba Suite ~\cite{chen2024video}& - &\Checkmark \\\cmidrule{3-5}
 &  & VideoMamba ~\cite{li2024videomamba}& \begin{tabular}[c]{@{}l@{}} FLOPs VideoMamba-Ti: 7.1 \\ VideoMamba-S: 28, VideoMamba-M: 83.1 \end{tabular}&\Checkmark \\\cmidrule{3-5}
 &  & SpikeMba ~\cite{li2024spikemba} & - &\XSolid \\\cmidrule{2-5}
 & \multirow{5}{2cm}{\centering Multi-Modal \\ understanding}& Cobra ~\cite{zhao2024cobra} &  -&\Checkmark \\\cmidrule{3-5}
 &   & ReMamber ~\cite{yang2024remamber} &  -&\XSolid \\\cmidrule{3-5}
 &  & VL-Mamba ~\cite{qiao2024vl}& - &\XSolid \\\cmidrule{2-5}
 & \multirow{4}{*}{Video prediction}  & VMRNN ~\cite{tang2024vmrnn} & Params 2.6, FLOPs 0.9 &\Checkmark \\\cmidrule{3-5}
& &HARMamba~\cite{li2024harmamba}& \begin{tabular}[c]{@{}l@{}} FLOPs 279.21 (PAMAP2), 237.83 (UCI) \\ 238.36 (UNIMIB HAR), 256.52 (WISDM) \end{tabular} & \XSolid \\
 \midrule
\multirow{34}{*}{Low-level vision} & Image super-resolution & MMA ~\cite{cheng2024activating}& - &\XSolid \\\cmidrule{2-5}
 & \multirow{5}{*}{Image restoration }& MambaIR ~\cite{guo2024mambair} & Params 16.7 &\Checkmark \\\cmidrule{3-5}
 &   & SERPENT ~\cite{shahab2024serpent} & - &\XSolid \\\cmidrule{3-5}
 &  & VmambaIR ~\cite{shi2024vmambair}& Params 10.50, FLOPs 20.5 &\Checkmark \\\cmidrule{2-5}
 & Image dehazing  & UVM-Net ~\cite{zheng2024u}& Params 19.25 &\Checkmark \\\cmidrule{2-5}
  & Image derain  & FreqMamba ~\cite{zhen2024freqmamba}& Params 14.52 &\XSolid \\\cmidrule{2-5}
 & Image deblurring  & ALGNet ~\cite{gao2024aggregating} & FLOPs 17 &\XSolid \\\cmidrule{2-5}
   &\multirow{7}{*}{Visual generation}  & MambaTalk ~\cite{xu2024mambatalk} &  -&\XSolid  \\\cmidrule{3-5}
  &   & Motion Mamba ~\cite{zhang2024motion}&   -&\Checkmark  \\\cmidrule{3-5}
  &  & DiS ~\cite{fei2024scalable}& - &\Checkmark  \\\cmidrule{3-5}
  &   & ZigMa ~\cite{hu2024zigma}& - &\Checkmark  \\\cmidrule{2-5}
  &\multirow{9}{*}{Point cloud}  & 3DMambaComplete ~\cite{li20243dmambacomplete} & 
  \begin{tabular}[c]{@{}l@{}}Params 34.06, FLOPs 7.12\end{tabular} 
  &\XSolid  \\\cmidrule{3-5}
  &   & 3DMambaIPF ~\cite{zhou20243dmambaipf}& -  &\XSolid  \\\cmidrule{3-5}
  &  & Point Cloud Mamba ~\cite{zhang2024point}&  
  \begin{tabular}[c]{@{}l@{}}Params 34.2, FLOPs 45.0 \end{tabular} 
  &\XSolid  \\\cmidrule{3-5}
  &   & POINT MAMBA ~\cite{liu2024point}&  Memory 8550&\Checkmark  \\\cmidrule{3-5}
   &   & SSPointMamba ~\cite{liang2024pointmamba}&  \begin{tabular}[c]{@{}l@{}}Params 12.3, FLOPs 3.6\end{tabular} &\Checkmark  \\\cmidrule{2-5}
 & 3D reconstruction & GAMBA ~\cite{shen2024gamba}   &  -&\XSolid \\\cmidrule{2-5}
 & Video generation  & SSM-based diffusion model ~\cite{oshima2024ssm}& -&\Checkmark \\
\bottomrule
\end{tabular}
\begin{tablenotes}
\item[1]  For the Efficiency, Inference speed is in ms, Memory is in MB, Params is in M, and FLOPS is in G.
\end{tablenotes}
\end{threeparttable}
\end{table}

\subsection{General Visual Mamba}

We categorize general vision-related tasks into High/Mid-level vision and Low-level vision. High/Mid-level vision encompasses recognition tasks such as classification, object detection, segmentation, and prediction across various input types, including images, videos, and 3D representation. In contrast, Low-level vision includes restoration, generation \etc.

\subsubsection{High/Mid-level vision}

The Visual Mamba Backbone~\cite{zhu2024vision,liu2024vmamba,yang2024plainmamba,huang2024localmamba,li2024mamba} performance decent on classification, object detection and segmentation. 
\emph{SSM-ViT} \cite{zubic2024state} is designed for object detection using event cameras. Unlike standard frame-based cameras, event cameras record per-pixel relative brightness changes in a scene as they occur. Therefore, object detection with event cameras requires processing an asynchronous stream of events in a four-dimensional spatio-temporal space.
Earlier studies have employed RNN architectures with convolutional or attention mechanisms to develop models that exhibit superior performance on downstream tasks using event camera data. However, these models often suffer from slow training. In response, the SSM-ViT block is introduced, leveraging SSM for efficient event-based information processing. It explores two strategies to mitigate aliasing effects when deploying the model at higher frequencies.

Leveraging Mamba's significant advancements in efficient training and inference with linear complexity, \emph{ReMamber} \cite{yang2024remamber} is introduced for referring image segmentation (RIS), a challenging task in the realm of multi-modal understanding. Distinguished from conventional segmentation, RIS entails identifying and segmenting specific objects in images based on textual descriptions. The ReMamber architecture comprises several Mamba Twister blocks, each featuring multiple VSS blocks and a Twisting layer. The Mamba Twister block serves as a multi-modal feature fusion block, receiving visual and textual features as input and outputting the fused multi-modal feature representation. Intermediate features are extracted after each Mamba Twister block and subsequently fed into a flexible decoder to generate the final segmentation mask. The VSS layers are tasked with extracting visual features, while the Twisting layer primarily captures effective visual-language interactions. The authors conduct experiments on multiple RIS datasets, achieving state-of-the-art results.Sigma~\cite{wan2024sigma} is a novel network for multimodal semantic segmentation tasks. One of them, Siam Mamba encoder, uses cascaded visual state space (VSS) blocks to extract multi-scale global information from different modalities. And an attention-based Mamba fusion mechanism and a channel-aware Mamba decoder are proposed. In the decoding stage, the fused features are further enhanced by channel-aware visual state space (CVSS) blocks, which can effectively capture multi-scale long-range information and realize cross-modal information integration.

Unlike transformers that rely on quadratic complexity attention mechanisms, Mamba, as a pure SSM-based model, excels in handling long sequences with linear complexity and proves particularly adept at processing lengthy videos at high resolutions. 
\emph{ViS4mer} \cite{islam2022long} serves as a model primarily used for recognizing and classifying long videos, particularly for understanding and categorizing lengthy movie clips. ViS4mer consists of two main components: a standard Transformer encoder designed for extracting short-distance spatiotemporal features from videos, and a multi-scale temporal S4 decoder optimized for subsequent long-range temporal reasoning. Leveraging the capability of the fundamental SSM to capture long-range dependencies in sequential data, the multi-scale temporal S4 decoder is implemented based on SSMs to reduce the computation cost of the model.

The \emph{Video Mamba Suite} \cite{chen2024video} is not a novel method; rather, it explores and assesses the potential of SSM, represented by Mamba, in video understanding tasks. The ViM block is enhanced into the Decomposed Bidirectionally Mamba (DBM) Block, which separates the input projector while sharing the parameters of SSM in both scanning directions. They classify Mamba into four distinct roles for modeling videos and compare it with existing Transformer-based models to evaluate its effectiveness in various video understanding tasks. Furthermore, the Video Mamba Suite comprises 14 models/modules to evaluate performance across 12 video understanding tasks. The experiments demonstrate that Mamba is applicable in video analysis and can be utilized for more complex, multimodal video understanding challenges.
Apart from the Video Mamba Suite, \emph{VideoMamba} \cite{li2024videomamba} is proposed for video understanding tasks, with a specific focus on addressing two major challenges: local redundancy and global dependencies. This study evaluates VideoMamba's capabilities across four key aspects: scalability in the video domain, sensitivity to short-term action recognition, advantages in long-term video understanding, and compatibility with other modalities.  To enhance model scalability in the visual domain, VideoMamba employs a self-distillation strategy. This approach significantly enhances VideoMamba's performance as both the model and input sizes increase, without the need for pre-training on large-scale datasets. While the ViM block enhances the model's spatial perception capabilities, VideoMamba extends this capability to 3D video understanding by including spatio-temporal bidirectional scanning. Through the extension of the ViM block, VideoMamba achieves a notable increase in processing speed and a decrease in computational resource consumption without compromising performance.SpikeMba~\cite{li2024spikemba} is a novel multimodal video content understanding framework designed to handle the task of temporal video localization in video content.SpikeMba combines Spiking Neural Networks (SNNs) and State Space Models (SSMs) to capture fine-grained relationships between multimodal input features. In particular, the Spike Saliency Detector (SSD) utilizes the thresholding mechanism of SNNs to generate saliency proposal sets that signal highly relevant or salient instances in the video through spikes. The Multimodal Relevance Mamba Block (MRM) is based on SSM and enhances the modeling of long range dependencies while maintaining linear complexity with respect to the input size.

In recent years, Multimodal Large Language Models (MLLMs) have been used with remarkable success in various domains. However, as a base model for many downstream tasks, current MLLMs consist of well-known Transformer networks with low secondary computational complexity. To improve the efficiency of these models, Cobra\cite{zhao2024cobra} has been proposed. This is a new multimodal large language model (MLLM) with linear computational complexity, which integrates the efficient Mamba language model into the visual modality, exploring different modal fusion schemes, and thus identifying ways to produce the most efficient multimodal representation.Cobra consists of three components: a visual encoder, a projector, and a Mamba backbone. Among them, the visual coder is used to extract the visual representation of the image, and the projector is used to transform the dimensions of the visual representation to match the dimensions of the tokens in the Mamba language model.The Mamba backbone, on the other hand, consists of a stack of 64 identical basic blocks that receive a combination of visual and textual embeddings while preserving the connectivity and the RMSNorm, and autoregressively transforms them into the target token sequences. VL-Mamba~\cite{qiao2024vl} is also a multimodal large language model consisting of a pre-trained visual coder, a randomly initialized MMC, and a pre-trained Mamba LLM. Among them, the visual coder uses the Vision Transformer (ViT) architecture to generate a sequence of patch features of the original image.For the MMC, a 2D visual selective scanning mechanism is proposed to solve the computer vision task since the state-space model is designed to deal with 1D sequential data with causality and the visual sequences generated by the visual coder are 2D non-causal data . The article explores three multimodal connector variants: the MLP (multilayer perceptron), the VSS-MLP (MLP combined with a VSS module) and the VSS-L2 (two linear layers combined with a VSS module). Input images are first acquired as visual features through a visual coder, then visual sequences are fed into the MMC, and finally the output vectors are fed into the Mamba LLM in combination with a tokenized text query in order to generate the corresponding response.The integration and processing of visual and verbal information is optimized through the synergy of these components.The Referential Image Segmentation (RIS) task requires a model to recognize and segment specific objects in an image based on textual descriptions, and ReMamber\cite{yang2024remamber} is a novel architecture for handling this task.ReMamber combines the Mamba model and introduces multimodal Mamba Twister blocks to explicitly simulate image-text interactions through its unique channeling and spatial warping mechanism Fusing textual and visual features.ReMamber extracts intermediate features after each Mamba Twister block and feeds them into a flexible decoder to generate the final segmentation mask.

To address the unparalleled challenge of predicting temporal and spatial dynamics for spatio-temporal forecasting in videos, the \emph{VMRNN} cell~\cite{tang2024vmrnn} introduces a novel recurrent unit designed to efficiently handle spatio-temporal prediction tasks. Recognizing the challenges in processing extensive global information, the VMRNN cell integrates VSS blocks with LSTM architecture to leverage the long-sequence modeling abilities of VSS blocks and the spatio-temporal representation capabilities of LSTM. 
This integration enhances the accuracy and efficiency of spatio-temporal predictions. The model conducts image-level analysis by segmenting each frame into patches, which are subsequently flattened and processed through an embedding layer. This process enables the VMRNN layer to extract and predict spatio-temporal features effectively.
HARMamba~\cite{li2024harmamba} builds on ViT blocks for activity recognition and reaches superior performance while reducing computational and memory overhead on the activity recognition tasks.

\subsubsection{Low-level vision}

In the realm of image super-resolution, Meet More Areas (\emph{MMA})~\cite{cheng2024activating} stands out as a novel model designed for super-resolution tasks. Built on the ViM block, MMA aims to enhance performance by activating a wider range of areas within images. To achieve this, MMA adopts several key strategies, including integrating ViM into MetaFormer-style modules, pre-training ViM on larger datasets, and employing complementary attention mechanisms. MMA comprises three main modules: shallow feature extraction, deep feature extraction, and high-quality reconstruction. Leveraging the ViM module, MMA effectively models global information and further expands the activation region through attention mechanisms.

Existing restoration backbones often face the dilemma between global receptive fields and efficient computation, hindering their application in practice, while Mamba shows great potential for long-range dependency modeling with linear complexity, which offers a way to resolve the above dilemma.
\emph{MambaIR}~\cite{guo2024mambair} aims to solve the problem by introducing local enhancement and channel attention mechanisms to improve the standard Mamba model. The methodology of the model mainly consists of three stages: shallow feature extraction, deep feature extraction, and high-quality image reconstruction. Among them, the deep feature extraction stage utilizes multiple residual state space blocks (RSSBs) for feature extraction, which add a VSS block before the channel attention block designed in previous transformer-based restoration networks.
\emph{SERPENT}~\cite{shahab2024serpent} designs a hierarchical architecture that processes input images in a multi-scale manner, including processing steps such as segmentation, embedding, downsampling, and upsampling, and introduces jump connections to facilitate information flow. Among them, the Serpent block is the main processing unit, consisting of multiple VSS blocks stacked on each other. Serpent combines the advantages of Convolutional Networks and Transformers to drastically reduce the computational effort, GPU memory requirement, and model size while maintaining high reconstruction quality.
\emph{VmambaIR}~\cite{shi2024vmambair} proposes the OSS module to comprehensively and efficiently model image features from six directions. The omnidirectional selective scanning mechanism overcomes the unidirectional modeling limitation of SSMs and achieves comprehensive pattern recognition and modeling by modeling the image information flow in all three dimensions.

\emph{UVM-Net}~\cite{zheng2024u} is a novel single-image defogging network architecture that combines the local feature extraction of convolutional layers and SSM's long-range dependency modeling capability to exhibit efficient performance. The method employs an encoder-decoder network architecture, and the critical component is the ViM block, which leverages the long-range modeling capability of SSM by rolling the feature map over the channel domain. Unlike U-Mamba~\cite{ma2024u} and Mamba-UNet ~\cite{wang2024mamba}, the ViM block establishes long-range dependencies on another dimension of the feature map (the non-channel domain).

Images lose important frequency information under the influence of raindrops, which affects the performance of visual perception and advanced visual tasks.FreqMamba~\cite{zhen2024freqmamba} is a novel image de-raining method that combines Mamba modeling and frequency analysis techniques to solve the image de-raining problem. Specifically, FreqMamba contains three branching structures, including spatial Mamba, frequency band Mamba, and Fourier global modeling. Spatial Mamba processes raw image features to extract details and correlations within the image. Frequency Band Mamba uses the Wavelet Packet Transform (WPT) to decompose the input features into spectral features in different frequency bands and scan them along the frequency dimension. Fourier modeling i.e., processing the input using the Fourier transform captures the global degradation patterns affecting the image. Extensive experiments have shown that FreqMamba outperforms existing state-of-the-art methods both visually and quantitatively.

Image deblurring is a classical problem in low-level computer vision, which aims to recover high-quality clear images from blurred input images.ALGNet~\cite{gao2024aggregating} is an efficient image deblurring network that utilizes selective state-space models (SSM) to aggregate rich and accurate features.The network consists of multiple ALGBlocks, each of which contains a CLGF module that captures local and global features and a feature aggregation module FA.The CLGF module captures long-range dependent features using SSM and employs a channel-attention mechanism to reduce local pixel forgetting and channel redundancy.The FA module emphasizes the importance of local features in recovery by recalibrating the weights.

The efficiency of Mamba contributes significantly to mitigating the high computational complexity associated with training generation tasks.
To address the change in generating long and diverse sequences with low latency, \emph{MambaTalk}~\cite{xu2024mambatalk} implements a two-stage modeling strategy with discrete motion priors to enhance the quality of gestures and employs mamba block to enhance gesture diversity and rhythm through multimodal integration.
\emph{Motion Mamba}~\cite{zhang2024motion} is introduced to construct a motion generation model based on Mamba, leveraging its efficient hardware-aware design. Motion Mamba consists of two main components: the Hierarchical Temporal Mamba (HTM) block for temporal data handling, and the Bidirectional Spatial Mamba (BSM) block for processing latent poses. The HTM block employs several isolated SSM modules within a symmetric U-Net architecture to maintain motion consistency across frames. Meanwhile, the BSM block enhances the accuracy of motion generation within a temporal frame by processing latent poses bidirectionally.
Diffusion State Space Models (\emph{DiS}) ~\cite{fei2024scalable} substitute the conventional U-Net backbone in diffusion models with SSMs. This framework considers all inputs, including time, conditions, and noisy image patches and tokens.
To tackle the oversight of spatial continuity in the scanning scheme of existing Mamba-based vision methods, Zigzag Mamba~\cite{hu2024zigma} is introduced as a straightforward, plug-and-play solution inspired by DiT style approaches. Essentially, it retains the scanning scheme of Plain Mamba but expands it from four to eight schemes by incorporating their mirror flipping schemes, as shown in (f) Fig.~\ref{fig:2dscanning}. Subsequently, Zigzag Mamba is integrated with the Stochastic Interpolant framework, forming \emph{ZigMa}, to explore the scalability of the diffusion model on large-resolution visual datasets.\emph{GAMBA}~\cite{shen2024gamba} introduces a sequential network based on Mamba, which enables context-dependent reasoning and linear scalability for sequence length. This architecture accommodates many Gaussians for the 3D Gaussian splatting process.
To address the issue of quadratic memory consumption increase with sequence length in traditional attention-based video generative diffusion models, \emph{SSM-based diffusion model}~\cite{oshima2024ssm} is introduced for generating longer video sequences. Like ViS4mer ~\cite{islam2022long}, the SSM-based diffusion model reimagines the attention modules within the conventional temporal layers of Video Diffusion Models (VDMs). It replaces them with a ViM block designed to capture the temporal dynamics of video data, accompanied by a Multi-Layer Perceptron (MLP) to boost model performance. This innovative approach significantly mitigates memory consumption for extended sequences.

The irregularity and sparsity of point cloud data have been a challenge in 3D vision. Although Transformer shows potential in point cloud analysis tasks based on its powerful global information modeling capability, its computational complexity grows significantly with the increase of input length, limiting its application to long sequence models. In this context, SSPoint Mamba~\cite{liang2024pointmamba} is proposed as a simple and effective state space model for point cloud analysis. The model uses embedded point blocks as inputs and enhances the global modeling capability of SSM with a reordering strategy that provides a more rational geometric scanning order. The reordered point tokens (point tokens) are fed into a series of Mamba blocks to causally capture the point cloud structure. The model demonstrates its effectiveness on several point cloud analysis tasks.3DMambaComplete~\cite{li20243dmambacomplete} network aims to address the computational complexity challenges posed by the loss of local details and attention mechanisms common in point cloud completion, and it is built on top of the novel Mamba framework. The method first downsamples incomplete point clouds, then enhances feature learning with Mamba Encoder, then predicts and refines hyperpoints, and disperses hyperpoints to different 3D locations by learning specific offsets, and finally performs point deformation to generate a complete point cloud. The model utilizes the concept of structured state-space modeling to improve shape reconstruction by predicting hyperpoints and point deformation to control the deformation at each hyperpoint location.The point cloud filtering model 3DMambaIPF~\cite{zhou20243dmambaipf}, on the other hand, aims to deal with the denoising of large-scale point cloud data. The network integrates Mamba into a filtering module, Mamba-Denoise, to achieve accurate and fast modeling of long sequences of point cloud features.3DMambaIPF consists of several Mamba-Denoise modules, and employs a strategy of iterative point cloud filtering to implement the filtering process for point clouds. The loss function includes reconstruction loss and differentiable rendering loss to minimize the distance between the noisy point cloud and the real point cloud, optimize the visual boundary of the point cloud, and improve the realism of the denoising results.Point Cloud Mamba~\cite{zhang2024point} combines local and global modeling frameworks and proposes a Consistent Traversal Serialization (CTS) approach to transform 3D point cloud data into 1D point sequences while ensuring that adjacent points in the sequence are also spatially adjacent. In addition, point cueing and position encoding based on spatial coordinate mapping are introduced to help Mamba process point sequences more efficiently and inject position information.Point Mamba~\cite{liu2024point} is a new backbone network for point cloud processing that addresses the causality of state space models on point cloud data by introducing an octree-based ordering strategy. In addition, Point Mamba blocks incorporate a bi-directional selective scanning mechanism to adjust the sequence order dependency of Mamba.

%% file: tex/conclu.tex
\section{Conclusion}

Mamba is gaining prominence in computer vision for its ability to manage long-range dependencies and its significant computational efficiency relative to Transformers. As detailed in recent surveys, various methods have been developed to harness and explore Mambas' capabilities, reflecting ongoing advancements in the field.

We begin by discussing the foundational concepts of SSMs and Mamba architectures, followed by a comprehensive analysis of various competing methodologies across a spectrum of computer vision applications. Our survey encompasses state-of-the-art Mamba models designed explicitly for backbone architectures, high/mid-level vision, low-level vision, medical imaging, and remote sensing. This survey is the first review paper about the recent developments in SSMs and Mamba-based techniques, explicitly focusing on computer vision challenges. Our goal is to generate more interest among the vision community in utilizing the possibilities of Mamba models and finding solutions to their current limitations.